\documentclass[11pt]{article}

\usepackage[preprint]{acl}

\usepackage{times}
\usepackage{latexsym}

\usepackage[T1]{fontenc}
\usepackage[utf8]{inputenc}

\usepackage{microtype}

\usepackage{inconsolata}

\usepackage{graphicx}

\usepackage{booktabs}
\usepackage{longtable}
\usepackage{multirow}
\usepackage{xcolor}
\usepackage{amsmath}
\usepackage{tikz}

\usepackage{tabularx}

\usepackage{amssymb}
\usepackage{placeins}

\usetikzlibrary{arrows.meta,shapes.geometric,shapes.misc}

\newcommand{\benchmarkname}{TwinICL}

\title{\benchmarkname: Diagnosing Multimodal In-Context Learning through \\ Paired Counterfactuals}

\author{
  Zihan Xue$^1$ $\:\:$ Po-Yi Lu$^{*,2}$ $\:\:$ Serhii Honcharenko$^{*,3}$ $\:\:$ Zih-Ching Chen$^4$ $\:\:$ Hsuan-Tien Lin$^2$ \\ 
  \bf $\:\:$ Nanyun Peng$^1$ $\:\:$ I-Hung Hsu$^5$ $\:\:$ Kuan-Hao Huang$^3$ \vspace{0.2em}\\
  $^1$University of California, Los Angeles $\:\:$
  $^2$National Taiwan University \\
  $^3$Texas A\&M University $\:\:$
  $^4$NVIDIA AI Technology Center $\:\:$
  $^5$Arena Intelligence Inc \vspace{0.2em}\\
  \texttt{zihanxue@ucla.edu}, $\:\:$
  \texttt{d09944015@csie.ntu.edu.tw}, $\:\:$
  \texttt{serhii@tamu.edu}, \\
  \texttt{ihung@arena.ai}, $\:\:$
  \texttt{khhuang@tamu.edu} \\
}

\newcommand\nnfootnote[1]{%
  \begin{NoHyper}
  \renewcommand\thefootnote{}\footnote{#1}%
  \addtocounter{footnote}{-1}%
  \end{NoHyper}
}

\begin{document}
\raggedbottom
\maketitle
\begin{abstract}
In-context learning (ICL) enables models to infer tasks from demonstrations, but existing benchmarks generally lack matched text and image versions needed to compare ICL performance across modalities.
We introduce \benchmarkname{}, a procedurally generated benchmark providing such pairs for controlled comparison.
Across six open-weight models and 38 tasks, multimodal ICL consistently underperforms text-only ICL, with gaps varying by task family.
To test whether this gap can be recovered, we target visual access, task framing, and reasoning through three interventions. Their combination recovers strong multimodal ICL performance on a diagnostic subset, despite limited or inconsistent individual effects.
To distinguish difficulties in executing tasks from those in inferring them, we evaluate models with explicit task instructions, revealing a modality gap even when the task is known.
We then examine how adding demonstration inputs and outputs reshapes this gap, highlighting demonstrations' dual role as additional context to process and evidence about the task.
The dataset is available at \url{https://github.com/lab-flair/TwinICL}.
\nnfootnote{$^{*}$ Joint second authorship. Zihan Xue, Po-Yi Lu, and Serhii Honcharenko all contributed to writing and revising the manuscript. See \hyperref[sec:author-contributions]{Author Contributions} for details.}
\end{abstract}

\section{Introduction}

\begin{figure*}[!t]
\centering
\includegraphics[width=1\textwidth]{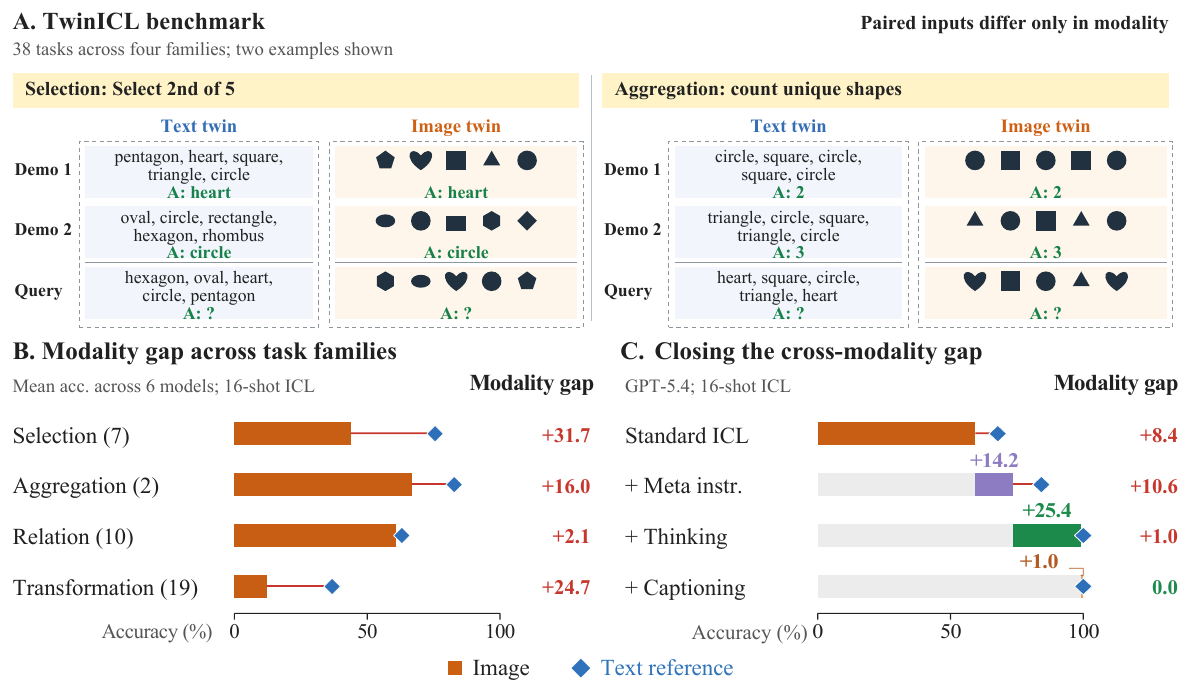}
\caption{
\benchmarkname{} compares paired text and image renderings of the same ICL episode. \textbf{(A)} Two example tasks from \benchmarkname{}'s 38 tasks across four families, each shown as a matched text/image pair with the same underlying examples and textual answers. Only the input modality changes. Task rules are shown for the reader and are not included in the model prompts.
\textbf{(B)} Mean 16-shot accuracy (\%) across six models, with equal weight per model and per task within each family. Parentheses give task counts; Table~\ref{tab:main-benchmark-results} reports per-model results. Bars show accuracy in the image condition (multimodal ICL), and diamonds show accuracy in the matched text condition (text-only ICL).
\textbf{(C)} Incremental ablation for GPT-5.4 at 16 shots, with accuracy (\%) averaged equally across five diagnostic tasks (Table~\ref{tab:prompting-results}). Rows cumulatively add meta instruction, thinking, and ground-truth shape-name captions. After the baseline, colored segments show the change in accuracy in the image condition from the previous row. Gap is accuracy in the text condition minus that in the image condition within the same row; both differences are measured in percentage points. The matched text condition uses the same meta-instruction and thinking settings; captions apply only to images.
}
\label{fig:twinicl-teaser}
\end{figure*}

In-context learning (ICL) allows a large language model to acquire a new task from a few demonstrations at inference time, without updating its parameters~\citep{brown2020language,dong2024survey}. Multimodal large language models (MLLMs) extend this ability to interleaved text and image inputs~\citep{tsimpoukelli2021multimodal,alayrac2022flamingo,zhao2024mmicl}.
This extension is especially valuable because demonstrations can communicate visual procedures directly, without requiring the user to describe every relevant object, relation, or transformation in language. It is therefore important to understand whether models can use visual demonstrations as reliably as textual ones.

Recent evaluations raise doubts about this reliability: models sometimes ignore the visual context or rely on textual cues and shallow heuristics~\citep{zong2025vlicl,chen2025truemicl,chen2025canmllms,huang2025mimicking,SG2025_arXiv_WhatVLMSee}.
These findings show that multimodal ICL can be brittle, but their interpretation remains ambiguous: The observed failures may arise from challenges introduced by multimodality, from the difficulty of the underlying ICL problem regardless of modality, or from both. Determining how much is attributable specifically to multimodality requires a matched reference: how would the same model solve the same ICL problem if it were presented in text?

We introduce \benchmarkname{}, a procedurally generated benchmark that provides this reference through paired text and image renderings. Each ICL episode is generated in a modality-independent form and then rendered as a text twin and an image twin that differ only in whether their inputs are presented as text or images; demonstration answers and targets remain textual in both. We call the resulting matched pair a \emph{paired counterfactual} because it presents the same underlying ICL problem with only the input modality changed (Figure~\ref{fig:twinicl-teaser}A). The performance difference between the twins, which we call the \emph{cross-modality performance gap} or modality gap, captures how performance changes when the same task evidence is presented visually rather than textually. Beyond enabling this paired comparison, procedural generation supports systematic variation in task properties while holding other factors fixed. \benchmarkname{} comprises 38 tasks grouped into four families: Selection, Relation, Aggregation, and Transformation. It serves both as a controlled benchmark of the modality gap and as a flexible testbed for studying when multimodal ICL succeeds or fails.

Our evaluation reveals a persistent modality gap: on average across tasks, presenting the same ICL problems visually lowers performance at every evaluated model--shot combination.
The magnitude of this gap varies substantially across task families, with especially pronounced differences in Selection and Transformation (Figure~\ref{fig:twinicl-teaser}B).
Moreover, models that perform strongly in text-only ICL can perform substantially worse in multimodal ICL, indicating that text-only ICL capability does not fully carry over to visual inputs.

To better understand the persistent modality gap and whether it can be overcome, we apply targeted interventions addressing visual access, task framing, and reasoning. On a focused subset of high-gap tasks, combining these interventions restores strong multimodal ICL performance (Figure~\ref{fig:twinicl-teaser}C), even though each intervention alone has limited or inconsistent effects. Complementing this analysis of recoverability, we examine how the gap should be interpreted. Multimodal ICL brings together two challenges: working with visual inputs and learning the task from demonstrations. Through controlled variants of the standard ICL setting, we find that the gap persists even when the task is stated directly, showing that standard multimodal ICL builds on a more basic difficulty in carrying out known tasks across text and images. Demonstrations then play a dual role: they introduce additional context that must be processed while also providing useful evidence about the shared task. Together, these findings show that the modality gap reflects the interaction of several capabilities that a single end-to-end score can obscure. By enabling controlled comparisons across these factors, \benchmarkname{} provides a more precise framework for analyzing and improving multimodal in-context learning.

In summary, we make three contributions.
(1) We introduce \benchmarkname{}, a procedurally generated benchmark built around paired counterfactuals: matched text and image renderings of the same ICL episode. This design isolates the effect of input modality while holding task content fixed.
(2) We use \benchmarkname{} to identify a persistent modality gap across six models. The gap varies substantially across task families, and strong text-only ICL performance does not reliably translate to multimodal ICL.
(3) We conduct complementary recovery and decomposition analyses of this gap. The recovery analysis shows that multimodal ICL performance can be substantially recovered when visual access, task framing, and reasoning support are combined. The decomposition analysis shows that the gap persists beyond task induction and is further shaped by multi-image context.

\section{Related Work}
\label{sec:related-work}

\subsection{Multimodal In-Context Learning}

Early multimodal systems demonstrated that language-model few-shot capabilities can extend to interleaved image--text inputs \citep{tsimpoukelli2021multimodal,alayrac2022flamingo,zhao2024mmicl}. Subsequent benchmarks examine the breadth and reliability of this ability. VL-ICL Bench evaluates diverse image-to-text and text-to-image ICL tasks \citep{zong2025vlicl}, while UniICL organizes multimodal ICL through a capability-oriented taxonomy and a broad evaluation suite \citep{xu2026uniicl}. Diagnostic studies report that models can rely heavily on textual demonstration content, underuse visual context, or exploit shallow heuristics rather than consistently infer the intended task \citep{chen2025canmllms,chen2025truemicl,SG2025_arXiv_WhatVLMSee,huang2025mimicking}. Complementary work improves demonstration selection and retrieval \citep{zhang2023goodexamples,zhou2024visualicl,yi2025drum}.

These evaluations motivate studying how task demands and input modality affect the use of demonstrations. Outside ICL, SEAM compares semantically equivalent textual and visual inputs to evaluate consistency across representations \citep{tang2025seam}; it does not study hidden-task induction from demonstrations. \benchmarkname{} contributes an extensible framework for constructing paired text--image ICL tasks and a benchmark spanning diverse task families with controlled task variations. This combination supports systematic investigation of how modality gaps vary across tasks and experimental conditions.

\subsection{Reasoning and Diagnosis in Multimodal ICL}

\citet{wang2026lagsbehind} construct paired textual and visual versions of an outlier-detection scenario to investigate task-mapping construction and transfer. Their analysis identifies difficulties in transferring demonstration-derived mappings to the query. Their controlled comparison centers on a specific scenario for mechanism investigation; \benchmarkname{} provides a broad benchmark and an extensible task-construction framework for systematic evaluation across task families and controlled variations.

\citet{wang2026inductivedeductive} identify an inductive gap in multimodal ICL and propose an inductive--deductive framework combining visual token compression, attention rebalancing, chain-of-thought guidance, and auxiliary training. Beyond ICL, Multimodal-CoT separates rationale generation from answer inference \citep{zhang2023multimodalcot}, while Visual CoT makes intermediate visual evidence explicit through grounded reasoning \citep{shao2024visualcot}. These approaches motivate examining how reasoning support affects multimodal task performance.

Our diagnostic experiments test how enabling native model thinking affects the modality gap, both alone and alongside support for visual access and task framing. Explicit-rule and context controls further examine whether the gap persists when the task is already specified and how demonstration inputs and outputs reshape it. These experiments characterize the conditions under which multimodal ICL performance recovers and investigate the demands contributing to the modality gap.

\section{\benchmarkname} \label{sec:benchmark}

\begin{figure*}[!t]
    \centering
    \includegraphics[width=0.8\textwidth]{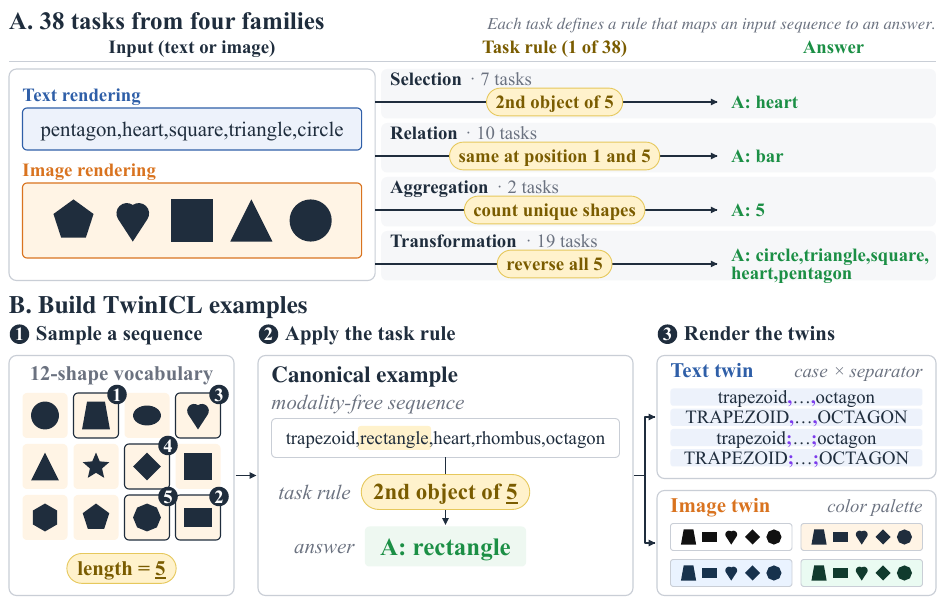}
    \caption{\benchmarkname{} tasks and example construction.
    \textbf{(A)} Each task defines a rule that maps an ordered shape sequence to a text answer. The 38 rules fall into four families; one rule per family is applied to the same input, shown in its text and image renderings.
    \textbf{(B)} A shape sequence is sampled from the 12-shape vocabulary (here five shapes) and the task rule computes its answer; this canonical example is fixed before any rendering.
    Its input is rendered in four text variants (case $\times$ separator)
    and four image variants (color palette), preserving shape identities
    and order. All variants share the same canonical target. 
    }
    \label{fig:twinicl-construction}
    \end{figure*}

    \benchmarkname{} is designed to compare the text and image conditions on the same underlying ICL problems.
    When tasks or examples differ across modalities, an accuracy gap can reflect those differences
    as well as the input representation.
    We therefore generate shared input-answer examples and use them to construct paired ICL
    prompts with the same task rule, demonstrations, query, and target answer content.
    We first introduce the tasks and their coverage (Figure~\ref{fig:twinicl-construction}A), then explain how canonical examples are generated and rendered (Figure~\ref{fig:twinicl-construction}B) and assembled into paired prompts (Figure~\ref{fig:twinicl-teaser}A).
    Finally, we report the benchmark's scale and integrity checks.

    \subsection{Tasks and Coverage} \label{sec:task-coverage}
        To construct matched text and image inputs, \benchmarkname{} uses a vocabulary of 12 named shapes,
        each represented by its name or its visual form (Figure~\ref{fig:twinicl-construction}A).
        Each task applies a hidden rule to an ordered sequence of these shapes.
        For example, \emph{Select 2nd of 5} returns ``heart'' for the input ``pentagon, heart, square, triangle, circle.''
        Table~\ref{tab:task-family-statistics} summarizes the 38 tasks across
        four families: Selection, Relation, Aggregation, and Transformation.
        Relation tasks return \texttt{foo} for even sequence lengths or matching
        shapes, and \texttt{bar} otherwise.
        Figure~\ref{fig:twinicl-construction}A illustrates these families,
        and Appendix~\ref{sec:appendix-task-taxonomy} provides the complete
        task definitions.

        \begin{table}[t]
            \centering
            \small
            \setlength{\tabcolsep}{3pt}
            \renewcommand{\arraystretch}{1.04}
        
            \begin{tabularx}{\columnwidth}{
                @{}
                >{\raggedright\arraybackslash}X
                r
                @{}
            }
                \toprule
                \textbf{Task group} & \textbf{Tasks} \\
                \midrule
        
                \multicolumn{2}{@{}l@{}}{\textbf{Selection (7 tasks)}} \\
                Positional selection$^{\dagger}$ (5 shapes)
                    & 5 \\
                Select most frequent shape (9 or 11 shapes)
                    & 2 \\
        
                \addlinespace[5pt]
                \multicolumn{2}{@{}l@{}}{\textbf{Relation (10 tasks)}} \\
                Test even/odd sequence length
                    & 1 \\
                Same/different: first and last (2, 5, or 7 shapes)
                    & 3 \\
                Adjacent same/different$^{\dagger}$ (7 shapes)
                    & 6 \\
        
                \addlinespace[5pt]
                \multicolumn{2}{@{}l@{}}{\textbf{Aggregation (2 tasks)}} \\
                Count all shapes
                    & 1 \\
                Count distinct shapes
                    & 1 \\
        
                \addlinespace[5pt]
                \multicolumn{2}{@{}l@{}}{\textbf{Transformation (19 tasks)}} \\
                Merge adjacent duplicates (5 or 7 shapes)
                    & 2 \\
                Reverse entire sequence (5 or 7 shapes)
                    & 2 \\
                Rotate left/right by 2 positions (5 or 7 shapes)
                    & 4 \\
                Window reversal$^{\dagger}$ (3 of 7 shapes)
                    & 5 \\
                Adjacent swap$^{\dagger}$ (7 shapes)
                    & 6 \\
        
                \bottomrule
            \end{tabularx}
        
            \caption{
                Coverage of the 38 tasks in \benchmarkname{}.
                Each fixed rule counts as a task.
                For example, in \emph{window reversal}, a seven-shape sequence
                has five three-shape windows: positions 1--3, 2--4, 3--5, 4--6, and 5--7.
                Reversing each window defines a separate task.
                Marked groups ($\dagger$) contain the 22 position-controlled tasks.
            }
            \label{tab:task-family-statistics}
        \end{table}

        To examine how accuracy and the modality gap vary across sequence
        positions, we organize 22 of the 38 tasks into four position-controlled
        groups: positional selection, adjacent same/different, window reversal,
        and adjacent swap. Within each group, the operation and sequence length
        remain fixed while the position at which the operation acts changes.
        The five positional selection tasks vary which shape is returned from
        a five-shape sequence. For seven-shape sequences, the six adjacent
        same/different tasks vary which neighboring pair is compared, the five
        window reversal tasks vary where a three-shape reversal window begins,
        and the six adjacent swap tasks vary which neighboring pair is exchanged.

    \subsection{Canonical Example Generation}
        For each task, we sample ordered shape sequences and compute their target answers
        by applying the task's rule (Figure~\ref{fig:twinicl-construction}B).
        The generator uses this rule, but it is not stated in the standard ICL prompts.
        Each resulting input-answer pair forms a canonical example, specified before either
        modality is rendered.
        We retain 132 unique canonical examples per task.

        Task-specific constraints exclude tied answers and inputs that obscure the operation.
        Positional selection and transformations that reorder shapes use 
        distinct shapes when repetitions would obscure the selected position or rearrangement.
        Most-frequent-item tasks reject ties for the highest frequency, 
        while adjacent-deduplication tasks require at least one pair of identical neighboring shapes.

    \subsection{Paired Rendering and Surface Variants}

        We render each input as shape names or an image of the corresponding shapes.
        For each task, we draw demonstrations and evaluation queries from
        two non-overlapping sets of canonical examples.
        An ICL episode consists of demonstrations and a query, with the query's
        target answer retained for evaluation.
        The resulting text and image prompts form a text twin and an image
        twin: both contain the same demonstrations in the same order and
        the same query, rendered in the corresponding modality
        (Figure~\ref{fig:twinicl-teaser}A).
        
        Each input has four text variants, combining lowercase or uppercase
        names with comma or semicolon separators, and four image variants
        using different color palettes on an $896 \times 896$ canvas
        (Figure~\ref{fig:twinicl-construction}B).
        Image layout remains fixed across palettes, and all variants preserve
        shape identities, order, and repetitions.
        Each prompt uses one variant consistently.
        
        Answers remain textual and share the same canonical target.
        For text inputs, answers follow the input's casing and separator.
        For image inputs, answers use lowercase shape names and relation
        labels, with comma-separated sequences.
        Counts are written as numerals.

    \subsection{Scale and Integrity Checks}

        Across its 38 tasks, \benchmarkname{} contains 5,016 canonical examples
        and 40,128 rendered examples, divided equally between text and image inputs.
        These totals describe the example pool from which we assemble complete
        ICL episodes.
        
        Shared canonical identifiers link each rendering to its underlying
        input-answer pair.
        They provide a common key for checking that paired prompts contain
        the same demonstrations in the same order and the same query.
        The shared identifiers also allow stored answers to be checked
        for agreement across variants, accounting for the casing and
        separator conventions above.

\section{Benchmarking MLLMs on \benchmarkname{}}
\label{sec:main-results}
By pairing text and image renderings of the same underlying ICL episode, \benchmarkname{} enables a controlled measurement of how input representation affects ICL performance.
We first test whether a modality gap appears consistently across models and shot counts, and examine how it evolves as more demonstrations are provided.
We then use controlled comparisons across tasks to identify the task demands that amplify the gap.
Together, these results establish a persistent modality gap, and Section~\ref{sec:closing-gap} then asks which targeted forms of support can recover multimodal ICL performance.

\begin{figure*}[!t]
\centering
\includegraphics[width=\textwidth]{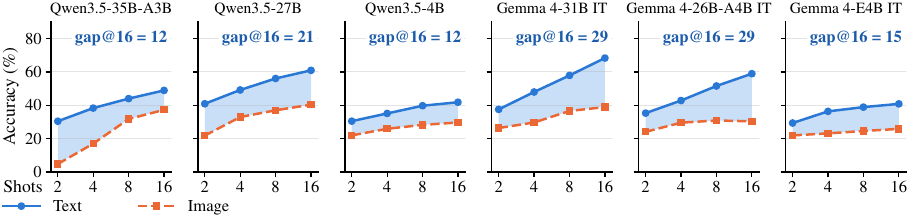}
\caption{Multimodal ICL underperforms text-only ICL at every evaluated shot count for all six models. Solid lines show accuracy (\%) in the text condition, and dashed lines show accuracy in the image condition; shaded bands show their difference, not uncertainty. Labels report the text-minus-image gap at 16 shots in percentage points. Each point averages the 38 tasks equally after averaging seeds and rendering variants within tasks.\textsuperscript{\ref{fn:evaluation-coverage}}
}
\label{fig:modality-gap}
\end{figure*}

\begin{table*}[!t]
\centering
\small
\setlength{\tabcolsep}{4pt}
\begin{tabular}{lcccccccccccc}
\toprule
\multirow{2}{*}{\textbf{Model}} & \multicolumn{3}{c}{\textbf{Selection}} & \multicolumn{3}{c}{\textbf{Relation}} & \multicolumn{3}{c}{\textbf{Aggregation}} & \multicolumn{3}{c}{\textbf{Transformation}} \\
\cmidrule(lr){2-4}\cmidrule(lr){5-7}\cmidrule(lr){8-10}\cmidrule(lr){11-13}
& \textbf{T} & \textbf{I} & \textbf{Gap} & \textbf{T} & \textbf{I} & \textbf{Gap} & \textbf{T} & \textbf{I} & \textbf{Gap} & \textbf{T} & \textbf{I} & \textbf{Gap} \\
\midrule
Qwen3.5-35B-A3B & 65.2 & 47.8 & \textbf{17.5} & 63.0 & 58.7 & 4.3 & 84.5 & 71.5 & 13.0 & 31.7 & 18.5 & 13.2 \\
Qwen3.5-27B & 74.2 & 51.6 & 22.6 & 62.3 & 61.0 & 1.3 & 89.8 & 73.3 & 16.5 & 52.3 & 21.8 & \textbf{30.5} \\
Qwen3.5-4B & 60.5 & 30.0 & \textbf{30.6} & 61.7 & 62.5 & $-$0.8 & 71.4 & 69.8 & 1.6 & 21.3 & 8.0 & 13.3 \\
Gemma 4-31B IT & 95.1 & 60.1 & 35.0 & 65.0 & 61.6 & 3.4 & 95.3 & 68.9 & 26.4 & 57.4 & 16.0 & \textbf{41.4} \\
Gemma 4-26B-A4B IT & 89.5 & 40.0 & \textbf{49.4} & 63.2 & 61.1 & 2.1 & 81.6 & 67.5 & 14.1 & 43.0 & 6.7 & 36.3 \\
Gemma 4-E4B IT & 68.9 & 33.9 & \textbf{35.0} & 63.2 & 61.2 & 2.1 & 74.0 & 49.5 & 24.5 & 15.1 & 1.7 & 13.4 \\
\bottomrule
\end{tabular}

\caption{Main 16-shot \benchmarkname{} results grouped by task family. \textbf{T} and \textbf{I} report ICL accuracy (\%) in the text and image conditions, respectively; \textbf{Gap} is T $-$ I within the same model and family, in percentage points. Positive gaps indicate lower accuracy in the image condition. Family scores average tasks equally after averaging available seeds and rendering variants within tasks.\textsuperscript{\ref{fn:evaluation-coverage}}
Bold Gap values mark the largest gap within each model.
}
\label{tab:main-benchmark-results}
\end{table*}

\subsection{Setup}
\label{sec:experimental-setup}
\paragraph{Models.}
We evaluate six open-weight MLLMs from two families: Qwen3.5 family and the Gemma~4 family, from small scale (4B) to large scale (31B or 35B).
All are instruction-tuned, and two are mixture-of-experts checkpoints.
Each model uses its own chat template and greedy decoding, so decoding is deterministic and involves no output-token sampling.
Hardware, API access, and decoding details are recorded in Appendix~\ref{sec:appendix-eval-details}.

\paragraph{Protocol and scoring.}

    We evaluate two matched conditions, illustrated in Figure~\ref{fig:twinicl-teaser}A.
    In the text condition (\textbf{T}), demonstration and query inputs are rendered as text;
    in the image condition (\textbf{I}), they are rendered as images.
    Demonstration outputs and model predictions remain textual in both conditions.
    We denote accuracy in these conditions by \textbf{T} and \textbf{I} and report the modality gap as \textbf{Gap} $=$ T $-$ I in percentage points. Positive values indicate lower accuracy in the image condition; negative values indicate higher accuracy.
Prompts contain only demonstrations followed by a query, without an instruction revealing the task.
We evaluate each task at 2, 4, 8, and 16 shots, and record 16 shots for task-level analysis in the main paper.
At every shot, the same seed selects the same canonical demonstrations and queries across the text and image conditions.

    The main benchmark protocol specifies 100 queries per task and condition,
    four rendering variants per modality, and three demonstration-sampling
    seeds.\footnote{\label{fn:evaluation-coverage}For current evaluation coverage, see Appendix~\ref{sec:evaluation-coverage}.}
    We average task accuracy over the available seeds and rendering
    variants, then give each task equal weight in family and overall
    averages.

    To avoid penalizing superficial output variations, we clean
    the extracted answers and targets before computing the exact-match accuracy.
    Cleaning lowercases the text, trims comma-separated tokens,
    discards empty tokens, and maps diamond to rhombus.
    A prediction is correct only when the cleaned answer matches the
    cleaned target in both order and multiplicity.
Appendix~\ref{sec:prompt-templates} provides the complete prompt formats, and Appendix~\ref{sec:appendix-eval-details} specifies the decoding settings, normalization procedure, and rendering variants.

\subsection{Benchmarking Results}
    Text inputs achieve higher task-averaged accuracy than image inputs at
    every evaluated model-shot combination (Figure~\ref{fig:modality-gap}).
    At 16 shots, the gap ranges from 12 to 29 percentage points across
    the six models and averages 20 points.
    Appendix~\ref{sec:additional-general-inference-tables} provides complete
    task-level results (Tables~\ref{tab:full-task-16shot-gap-selection}--\ref{tab:full-task-16shot-gap-transformation-swaps})
    and results at all evaluated shot counts (Table~\ref{tab:lower-shot-overall-gap}).

In the following paragraphs, we examine the main 16-shot results by task family and across the position-controlled tasks to identify which operations and task demands drive this gap.

\paragraph{The modality gap varies across task families.}
    At 16 shots Selection and Transformation show the largest modality gaps,
    ranging across models from 17.5 to 49.4 and from 13.2 to 41.4 
    percentage points, respectively (Table~\ref{tab:main-benchmark-results}).
    Aggregation gaps range from 1.6 to 26.4 points.
    Relation tasks have smaller gaps, but this does not necessarily
    indicate strong image performance:
    the six adjacent same/different tasks remain near chance in both modalities.
    Family-level gaps must therefore be interpreted alongside the underlying
    accuracies in the text and image conditions.

\begin{figure}[!t]
\centering
\includegraphics[width=\linewidth]{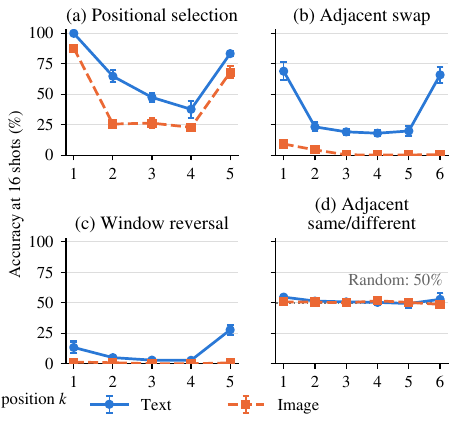}
\caption{Position profiles of text-only and multimodal ICL at 16 shots across the 22 position-controlled tasks. Solid and dashed lines show accuracy (\%) across positions in the text and image conditions, respectively.
Panels show positional selection, adjacent swap, window reversal, and adjacent same/different. Position $k$ identifies the selected shape in panel (a), the first shape of the pair $(k,k+1)$ in panels (b) and (d), and the start of the three-shape window in panel (c).
The dotted line in the adjacent same/different panel marks the 50\% random-guess baseline.
Each point averages rendering variants within each model and seed, then
the six evaluated models and three seeds equally.\textsuperscript{\ref{fn:evaluation-coverage}}
Error bars show one standard deviation across the three seed-level
means, each averaged over six models.}
\label{fig:position-sensitivity}
\end{figure}

\paragraph{Position sensitivity across modalities.}
    Figure~\ref{fig:position-sensitivity} shows how accuracy varies across
    positions for the four position-controlled groups across the six models.
    Positional selection retains endpoint advantages, while accuracy on image-based
    adjacent swaps remains near zero.
    Window reversals remain near zero in both modalities, and adjacent 
    same/different judgments stay near chance.

    Across the evaluated models and shot counts, image inputs yield lower
    task-averaged accuracy than their text counterparts.
    The gap varies across tasks and task families,
    and the position-controlled tasks show that a small gap can coexist with poor
    performance in both modalities.
    Section~\ref{sec:closing-gap} builds on these observations
    by testing which forms of support recover image
    performance on a diagnostic subset and how their effects interact.

\section{
    Diagnosing Multimodal ICL through Targeted Interventions
}
\label{sec:closing-gap}

Section~\ref{sec:main-results} establishes a persistent modality gap through controlled comparisons.
To move from measuring the gap toward understanding it, we test whether multimodal ICL performance can be recovered through targeted interventions during inference and what conditions enable that recovery.
When an intervention restores multimodal ICL performance, it identifies conditions under which the observed failure can be overcome. Comparing its effects across modalities and in combination with other interventions further reveals whether the benefit is image-specific and how different interventions interact.

\begin{table*}[!t]
    \centering
    \small
    \setlength{\tabcolsep}{3pt}
    \renewcommand{\arraystretch}{1.10}

    \begin{tabular*}{\textwidth}{
        @{\extracolsep{\fill}}
        l rrrr rrrr
        @{}
    }
        \toprule
        & \multicolumn{4}{c}{\textbf{GPT-5.4}}
        & \multicolumn{4}{c}{\textbf{Qwen3.5-35B-A3B}} \\
        \cmidrule(lr){2-5}
        \cmidrule(lr){6-9}
        \textbf{Incremental treatment}
            & \textbf{Text} & \textbf{Image}
            & \textbf{$\Delta$ Image} & \textbf{Gap}
            & \textbf{Text} & \textbf{Image}
            & \textbf{$\Delta$ Image} & \textbf{Gap} \\
        \midrule
        
        Standard ICL
            & 67.8 & 59.4 & -- & +8.4
            & 64.8 & 38.2 & -- & +26.6 \\
        \addlinespace[3pt]
        
        \quad + Meta instruction
            & 84.2 & 73.6 & +14.2 & +10.6
            & 65.8 & 34.2 & $-$4.0 & +31.6 \\
        
        \quad + Thinking
            & 100.0 & 99.0 & +25.4 & +1.0
            &  94.6 & 61.8 & +27.6 & +32.8 \\
        
        \quad + Caption
            & 100.0 & \textbf{100.0} & +1.0  & 0.0
            &  94.6 & \textbf{97.0}  & +35.2 & $-$2.4 \\
        \bottomrule
    \end{tabular*}

    \caption{
        Incremental ablation of meta instruction, thinking, and captioning
        on five diagnostic tasks at 16 shots.
        Text and Image report accuracy (\%) in the respective input conditions, averaged equally across tasks.
        Beginning from Standard ICL, each subsequent row cumulatively
        adds the intervention named in the first column.
        $\Delta$ Image is the Image value in the current row minus that
        in the previous row; Gap is Text minus Image within the same row.
        Both differences are measured in percentage points.
        Text uses the same meta-instruction and thinking settings;
        captions apply only to images.
        Negative gaps indicate an image-side advantage.
        All combinations and task-level scores appear in
        Tables~\ref{tab:intervention-details-gpt}
        and~\ref{tab:intervention-details-qwen}
        in Appendix~\ref{sec:appendix-intervention-details}.
    }
    \label{tab:prompting-results}
\end{table*}

\subsection{Interventions}
We organize the interventions around three broad demands of multimodal ICL: (1) accessing the relevant information in the images, (2) treating the demonstrations as evidence for a shared task, and (3) carrying out the inference needed to apply that task to the query.
For each demand, we introduce an intervention and measure whether it helps recover multimodal ICL performance: (1) ground-truth captions make the visual information directly available in text, (2) a meta instruction clarifies how the demonstrations should be used without revealing their rule, and (3) explicit thinking provides additional inference-time computation over the demonstrations and query.
\label{sec:interventions}
\paragraph{Captioning.}
We pair every demonstration and query image with a ground-truth textual caption consisting of its shape names in lowercase, separated by commas.
The caption immediately follows its image in the same user turn.

\paragraph{Meta instruction.}
In the standard-ICL condition, the model receives the 16 input--output demonstrations followed by the query, with no instruction explaining how to use the examples.
The meta-instruction condition adds a system instruction directing the model to infer the rule shared by the demonstrations and apply it to the query.
The instruction clarifies the role of the demonstrations without revealing the rule itself.
The exact instruction is found in Appendix~\ref{sec:prompt-templates}.

\paragraph{Explicit thinking.}
We enable each model's native thinking mode while leaving the prompt, demonstrations, and query unchanged.

\subsection{Individual Effects Vary, but Combined Interventions Recover Performance}

We conduct this diagnostic analysis on five tasks with pronounced
16-shot modality gaps.
We evaluate one closed model, GPT-5.4, and one open-weight model,
Qwen3.5-35B-A3B, across paired text and image inputs at 16 shots, using
a fixed rendering variant within each modality and 100 queries per
task and condition.
Appendix~\ref{sec:appendix-eval-details} provides the sampling and
decoding settings, while Tables~\ref{tab:intervention-details-gpt}
and~\ref{tab:intervention-details-qwen} report the selected tasks and
their complete task-level results.

Table~\ref{tab:prompting-results} shows one cumulative path through the
three interventions. We use accuracy in the image condition to measure performance recovery and the difference from the corresponding text condition to measure the remaining modality gap.

\paragraph{The full combination recovers strong image performance.}
When all three interventions are combined, both models achieve
near-ceiling accuracy in the image condition and no longer underperform their
corresponding text conditions (Table~\ref{tab:prompting-results}).
This result shows that strong multimodal ICL performance can be recovered on this diagnostic subset under targeted inference-time interventions.

\paragraph{Combined interventions can reverse individual effects.}
The interventions behave differently when combined than when used
individually. For Qwen3.5-35B-A3B, meta instruction and thinking each reduce image
performance individually, but increase it when combined
(Table~\ref{tab:intervention-details-qwen}). For GPT-5.4, thinking alone
produces strong image gains but substantial text regressions on some
tasks; combining it with meta instruction preserves the image gains
while reversing those regressions
(Table~\ref{tab:intervention-details-gpt}). Across the two models, these
results show that combined recovery reflects interactions that can
change the direction of individual effects, rather than a simple
accumulation of independent gains.

Section~\ref{sec:instruction-context-controls} next asks whether task induction is necessary for the gap and how multi-image context affects it.

\section{Disentangling the Modality Gap: Task Induction and Multi-Image Context}
\label{sec:instruction-context-controls}

\begin{table*}[!t]
\centering
\small
\setlength{\tabcolsep}{4pt}
\renewcommand{\arraystretch}{1.12}
\begin{tabular}{@{}l c p{4.2cm} c p{5.8cm}@{}}
\toprule
\textbf{Condition} & \begin{tabular}[t]{@{}c@{}}\textbf{GT}\\\textbf{instruction?}\end{tabular} & \textbf{Prompt schematic} & \begin{tabular}[t]{@{}c@{}}\textbf{Infer}\\\textbf{task?}\end{tabular} & \textbf{Question addressed} \\
\cmidrule(lr){1-5}
\textbf{0-shot + GT}
& Yes
& GT rule $\rightarrow$ query $\rightarrow$ answer
& No
& Does a modality gap remain when the task is stated and only the query is present? \\
\cmidrule(lr){1-5}
\textbf{Multi-input + GT}
& Yes
& GT rule $\rightarrow$ inputs$_{1:16}$ (no outputs) $\rightarrow$ query $\rightarrow$ answer
& No
& How does adding the input load of 16 demonstrations change that gap? \\
\cmidrule(lr){1-5}
\textbf{16-shot + GT}
& Yes
& GT rule $\rightarrow$ (input, output)$_{1:16}$ $\rightarrow$ query $\rightarrow$ answer
& No
& How do demonstration outputs change the effect of the same input context? \\
\cmidrule(lr){1-5}
\textbf{Regular 16-shot}
& No
& (input, output)$_{1:16}$ $\rightarrow$ query $\rightarrow$ answer
& Yes
& What changes when the task must be inferred rather than explicitly stated? \\
\bottomrule
\end{tabular}
\caption{Instruction and context control setups. GT denotes a ground-truth instruction that states the task rule but not the query answer. Each input is rendered as either a text sequence or its paired image, and all conditions use the same paired query. Where demonstrations are present, the sampled examples are also held fixed.}
\label{tab:instruction-context-setups}
\end{table*}

\begin{table*}[!t]
\centering
\footnotesize
\setlength{\tabcolsep}{4pt}
\renewcommand{\arraystretch}{1.08}
\begin{tabular}{l rrr rrr rrr rrr}
\toprule
\multirow{2}{*}{\textbf{Task}}
& \multicolumn{3}{c}{\textbf{0-shot + GT}}
& \multicolumn{3}{c}{\textbf{Multi-input + GT}}
& \multicolumn{3}{c}{\textbf{16-shot + GT}}
& \multicolumn{3}{c}{\textbf{Regular 16-shot}} \\
\cmidrule(lr){2-4}
\cmidrule(lr){5-7}
\cmidrule(lr){8-10}
\cmidrule(lr){11-13}
& \textbf{T} & \textbf{I} & \textbf{Gap}
& \textbf{T} & \textbf{I} & \textbf{Gap}
& \textbf{T} & \textbf{I} & \textbf{Gap}
& \textbf{T} & \textbf{I} & \textbf{Gap} \\
\midrule
\multicolumn{13}{l}{\textit{GPT-5.4}} \\
Select most frequent of 11
& 100.0 & 91.0 & 9.0 & 100.0 & 89.0 & \textbf{11.0} & 100.0 & 94.0 & 6.0 & 100.0 & 80.0 & 20.0 \\
Merge adjacent duplicates in 7
& 100.0 & 69.0 & 31.0 & 100.0 & 57.0 & \textbf{43.0} & 100.0 & 89.0 & 11.0 & 99.0 & 82.0 & 17.0 \\
Reverse 7
& 100.0 & 72.0 & 28.0 & 100.0 & 52.0 & \textbf{48.0} & 100.0 & 88.0 & 12.0 & 100.0 & 96.0 & 4.0 \\
Rotate left by 2 in 7
& 100.0 & 83.0 & 17.0 & 100.0 & 38.0 & \textbf{62.0} & 100.0 & 95.0 & 5.0 & 13.0 & 19.0 & $-$6.0 \\
Select 2nd of 5
& 100.0 & 90.0 & 10.0 & 100.0 & 93.0 & \textbf{7.0} & 100.0 & 100.0 & 0.0 & 27.0 & 20.0 & 7.0 \\
\midrule
\textbf{Average}
& \textbf{100.0} & \textbf{81.0} & \textbf{19.0} & \textbf{100.0} & \textbf{65.8} & \textbf{34.2} & \textbf{100.0} & \textbf{93.2} & \textbf{6.8} & \textbf{67.8} & \textbf{59.4} & \textbf{8.4} \\
\midrule
\multicolumn{13}{l}{\textit{Qwen3.5-35B-A3B}} \\
Select most frequent of 11
& 100.0 & 74.0 & 26.0 & 99.0 & 78.0 & \textbf{21.0} & 97.0 & 69.0 & 28.0 & 96.0 & 17.0 & 79.0 \\
Merge adjacent duplicates in 7
& 88.0 & 21.0 & 67.0 & 87.0 & 11.0 & \textbf{76.0} & 92.0 & 41.0 & 51.0 & 54.0 & 46.0 & 8.0 \\
Reverse 7
& 100.0 & 32.0 & 68.0 & 100.0 & 60.0 & \textbf{40.0} & 100.0 & 93.0 & 7.0 & 95.0 & 93.0 & 2.0 \\
Rotate left by 2 in 7
& 100.0 & 20.0 & 80.0 & 100.0 & 36.0 & \textbf{64.0} & 100.0 & 57.0 & 43.0 & 28.0 & 4.0 & 24.0 \\
Select 2nd of 5
& 100.0 & 80.0 & 20.0 & 99.0 & 88.0 & \textbf{11.0} & 100.0 & 92.0 & 8.0 & 51.0 & 31.0 & 20.0 \\
\midrule
\textbf{Average}
& \textbf{97.6} & \textbf{45.4} & \textbf{52.2} & \textbf{97.0} & \textbf{54.6} & \textbf{42.4} & \textbf{97.8} & \textbf{70.4} & \textbf{27.4} & \textbf{64.8} & \textbf{38.2} & \textbf{26.6} \\
\bottomrule
\end{tabular}%
\caption{Instruction and context controls without thinking on five diagnostic tasks. \textbf{T} and \textbf{I} report accuracy (\%) in the text and image conditions, respectively; \textbf{Gap} is T $-$ I within the same row and condition, in percentage points. Positive gaps indicate lower accuracy in the image condition. Average rows give each task equal weight. GT denotes a ground-truth task instruction; Table~\ref{tab:instruction-context-setups} defines the conditions. Boldface highlights the multi-input gaps.
}
\label{tab:instruction-context-controls}
\end{table*}
A modality gap measured under ICL is not necessarily specific to learning from demonstrations. In the standard benchmark setting, the model must determine the task from the examples and then perform that task on a new text or image input. The observed gap may therefore include a text--image difference that remains even when the model does not have to discover the task. Establishing whether such a difference exists is necessary for interpreting the benchmark: if it persists when the task is stated explicitly, multimodal task induction cannot be its sole source.

We therefore use instruction and context controls to ask two related questions: whether the modality gap persists when the task does not have to be learned from demonstrations, and how the gap changes when demonstration context is introduced around an explicitly stated task. The first tests whether task induction is the sole source of the gap; the second tests whether the amount and composition of the preceding context further alter it.

\subsection{Experimental Design}
We use four matched prompt conditions, summarized in Table~\ref{tab:instruction-context-setups}. Together, they address three questions: whether a modality gap exists when the task is already known, how adding multiple unlabeled inputs changes that gap, and whether labeled demonstrations make the same input context more useful. Comparing the explicitly instructed 16-shot condition with regular 16-shot ICL then tests how requiring task induction changes performance.

\textbf{0-shot + GT} provides the ground-truth (GT) task instruction and the query only. \textbf{Multi-input + GT} adds the 16 demonstration inputs but withholds their outputs, matching the input load and modality of standard 16-shot ICL without requiring task induction. \textbf{16-shot + GT} restores the corresponding outputs, allowing us to test whether labeled demonstrations make that same context more useful. Finally, \textbf{Regular 16-shot} keeps the labeled demonstrations and query but removes the explicit instruction. Comparing these last two conditions tests how requiring the model to infer the task from demonstrations changes performance: the examples are held fixed, and the task rule changes from stated to hidden.

We evaluate the same two models and five-task diagnostic subset as Section~\ref{sec:closing-gap}, using a fixed rendering variant within each modality and 100 queries per task and condition. Sampling and decoding settings are provided in Appendix~\ref{sec:appendix-eval-details}. The GT instructions are identical across models and modalities and are reported in Appendix~\ref{sec:prompt-templates}. We apply every condition to paired text (\textbf{T}) and image (\textbf{I}) renderings of the same query inputs and, where present, demonstration inputs. Reporting both accuracies and their signed difference, \textbf{Gap} $=$ T $-$ I, distinguishes modality-specific changes from those that affect both modalities. Because Section~\ref{sec:closing-gap} shows that thinking interacts with prompt design, the main analysis holds the reasoning regime fixed without thinking. Appendix~\ref{sec:appendix-instruction-context-thinking} reports the corresponding controls with thinking enabled.

\subsection{Results}
Table~\ref{tab:instruction-context-controls} reports the results for the four control conditions.
\paragraph{Task induction is not the only source of the modality gap.}
Under \textbf{0-shot + GT}, all 10 model--task pairs show a positive modality gap even though the prompt contains only the query and states the task explicitly. Thus, the gap cannot be attributed solely to task induction or multi-input context in these settings. This residual gap reflects image-side processes that remain when the task is explicitly stated: identifying the shapes in the image, applying the stated operation, and emitting the answer in the required format. Comparing \textbf{0-shot + GT} with \textbf{Multi-input + GT} next tests whether adding the input load of standard ICL introduces a further cost.

\paragraph{Unlabeled multi-image context compounds GPT-5.4's gap, but not Qwen3.5-35B-A3B's.}
Moving from \textbf{0-shot + GT} to \textbf{Multi-input + GT} restores the 16 demonstration inputs without their outputs while keeping the task explicit. For GPT-5.4, accuracy in the text condition remains perfect, but accuracy in the image condition falls and the gap widens on four of five tasks. This shows that additional image context can impose a cost even when the task is explicitly provided. Qwen3.5-35B-A3B shows no corresponding average accuracy decrease: its accuracy in the image condition improves, and the gap widens on only one task. Thus, unlabeled multi-image context creates a substantial additional burden for GPT-5.4 in this subset, but the effect does not generalize across the two models. Comparing \textbf{Multi-input + GT} with \textbf{16-shot + GT} next tests whether pairing those inputs with outputs changes that burden.

\paragraph{Output labels make multi-image context substantially more useful.}
Moving from \textbf{Multi-input + GT} to \textbf{16-shot + GT} restores the corresponding outputs. Accuracy in the image condition improves on all five GPT-5.4 tasks and four of five Qwen3.5-35B-A3B tasks, substantially narrowing the average gap for both models. Outputs therefore make the same multi-input context substantially more useful, but this comparison does not identify why: they may help bind each contextual input to its result, provide redundant evidence about the stated rule, or both. With complete labeled demonstrations restored, \textbf{16-shot + GT} becomes the closest explicitly instructed counterpart to the \textbf{Regular 16-shot} baseline.

\paragraph{Task induction lowers absolute accuracy and alters the existing modality gap.}
Moving from \textbf{16-shot + GT} to \textbf{Regular 16-shot} returns to the standard ICL baseline by removing the explicit task instruction. The model must therefore infer the task from the demonstrations, so standard ICL now requires task induction. Absolute accuracy falls sharply in both modalities for both models, while the modality gap shifts in both directions across the five tasks. The \textbf{0-shot + GT} results already establish that the gap can arise without learning from demonstrations. Returning to standard ICL adds task induction, which can amplify or reduce the modality gap depending on the task.

\paragraph{The modality gap reflects more than task induction.}
Together, these controls show that task induction is not the only source of the gap observed under standard ICL. A gap remains when the task is explicitly stated and only the query is presented; adding unlabeled demonstration inputs can widen it in a model-dependent manner; and pairing those inputs with outputs generally makes the same context more useful. Demonstrations thus play a dual role: they provide evidence about the shared task while also introducing multiple inputs that the model must process and relate.\footnote{The explicit-instruction, query-only gap also remains positive for all 10 model--task pairs with thinking enabled; see Appendix~\ref{sec:appendix-instruction-context-thinking}.} The modality gap should therefore be understood as the combined outcome of processing individual images, organizing multi-image context, inferring the shared task, and executing it---not as a direct measure of any single capability.

\section{Conclusion}
\label{sec:conclusion}
We introduced \benchmarkname{}, a procedurally generated benchmark that pairs text and image renderings of the same ICL episodes, enabling controlled measurement of the effect of input modality. Across six models and 38 tasks, multimodal ICL consistently underperforms text-only ICL. The gap varies substantially across task families, and strong text-only ICL performance does not reliably transfer to visually presented task evidence.

Our analyses show that multimodal ICL performance reflects interacting demands rather than a single bottleneck. The modality gap persists even when the task is known, while coordinated support for visual access, task framing, and reasoning can recover strong performance. Demonstrations serve both as evidence about the task and as additional multimodal context to process. By enabling controlled comparisons across modalities and task factors, \benchmarkname{} supports studying these capabilities both separately and, importantly, in interaction.

\section*{Author Contributions}
\phantomsection
\label{sec:author-contributions}
Zihan Xue led the project, designed and developed the benchmark dataset, designed and implemented analyses, and conducted experimental evaluations.
Po-Yi Lu designed and implemented analyses and conducted experimental evaluations.
Serhii Honcharenko conducted experimental evaluations.
Zihan Xue, Po-Yi Lu, and Serhii Honcharenko all contributed to writing and revising the manuscript.

\section*{Acknowledgment}
This research is partially supported by the Advanced Cyberinfrastructure Coordination Ecosystem: Services \& Support (ACCESS) program (No. CIS251227). Portions of this research were conducted with the advanced computing resources provided by Texas A\&M High Performance Research Computing.

\section*{Limitations}
\benchmarkname{} is primarily a benchmark for controlled comparisons of text-only and multimodal in-context learning, and its design deliberately trades naturalistic coverage for experimental control. The 12-shape vocabulary, ordered sequence tasks, and clean image renderings enable verifiable text--image pairing and systematic variation of task properties, while textual answers provide a shared output space across input modalities. This controlled setting allows performance differences to be studied with the underlying task content held fixed. The resulting findings are grounded in this domain; their transfer to natural images, cluttered scenes, and richer spatial tasks remains an empirical question. Extending paired construction to such settings while preserving meaningful control is an important direction for future work.

The accompanying diagnostic experiments investigate the modality gap through controlled changes at inference time. Interventions targeting visual access, task framing, and reasoning establish conditions for recovering strong performance, while instruction and context controls show that the gap extends beyond task induction alone. These analyses characterize model behavior under different forms of support; they do not establish how architecture and training give rise to the observed differences across modalities. Explaining these origins, and determining how the findings can inform training methods or architectural changes that improve models' use of visual demonstrations, remain directions for future work. \benchmarkname{} provides a controlled testbed for developing and evaluating such advances.

\bibliography{custom}

\clearpage
\appendix
\raggedbottom

\renewcommand{\dbltopfraction}{0.95}
\renewcommand{\textfraction}{0.05}
\renewcommand{\dblfloatpagefraction}{0.85}
\setcounter{dbltopnumber}{3}

\FloatBarrier
\section{Prompt Templates}
\label{sec:prompt-templates}

\paragraph{Prompt format for the open-weight models.}

Each ICL episode is formatted using the model's own chat template as alternating user--assistant turns. Every demonstration contributes a user turn containing either the rendered image in the image conditions or the input string in the text condition, followed by an assistant turn containing its output string. In the \textbf{I+Cap} condition, the caption follows its image inside the same user turn; the caption is the text rendering of the same canonical example's input, i.e., the shape names in lowercase separated by commas. The query forms the final user turn, and the model's completion is taken as its answer. A text prompt for Select 2nd of 5 therefore has the following structure (shortened to two demonstrations):

\begin{quote}\small\ttfamily
user: rhombus, oval, triangle,\\
\hspace*{1em}square, hexagon\\
assistant: oval\\
user: oval, circle, rectangle,\\
\hspace*{1em}hexagon, rhombus\\
assistant: circle\\
user: hexagon, oval, heart,\\
\hspace*{1em}circle, pentagon
\end{quote}

\noindent In the image conditions, each input string is replaced by the corresponding rendered image.

\paragraph{Prompt format for GPT-5.4.}

GPT-5.4 receives the ICL episode formatted as a single user message whose content interleaves text and image parts in the pattern \texttt{Q:}~\emph{input}~\texttt{A:}~\emph{output} for each demonstration, ending with \texttt{Q:}~\emph{query input}~\texttt{A:} so that the model completes the final answer. Captions, when present, follow their image exactly as in the open-weight format.

\paragraph{Meta instruction.}

The meta-instruction condition in Section~\ref{sec:closing-gap} prepends the following system-level instruction, identical for every model:
\begin{quote}\small\ttfamily
Learn the task from the demonstrations, then answer the final query with only the answer.
\end{quote}

\paragraph{Ground-truth task instructions.}

The instruction controls in Section~\ref{sec:instruction-context-controls} replace ICL task induction with a ground-truth instruction that states the task explicitly. The instruction is supplied at the system level. The five instructions are:
\begin{itemize}
\item \textbf{Select most frequent of 11:} ``Given an input sequence of 11 shapes, identify the shape that appears most frequently. Reply with only the lowercase shape name.''
\item \textbf{Merge adjacent duplicates in 7:} ``Given an input sequence of 7 shapes, merge adjacent duplicates by keeping one copy from each consecutive run of the same shape, preserving the remaining order. Reply with only the resulting lowercase shape names separated by commas.''
\item \textbf{Reverse 7:} ``Given an input sequence of 7 shapes, output the same shapes in reverse order. Reply with only the resulting lowercase shape names separated by commas.''
\item \textbf{Rotate left by 2 in 7:} ``Given an input sequence of 7 shapes, rotate the sequence left by 2 positions, moving the first 2 shapes to the end in the same order. Reply with only the resulting lowercase shape names separated by commas.''
\item \textbf{Select 2nd of 5:} ``Given an input sequence of 5 shapes, output the second shape in the sequence. Reply with only the lowercase shape name.''
\end{itemize}

\FloatBarrier
\section{Inference and Decoding Details}
\label{sec:appendix-eval-details}

\paragraph{Answer normalization and scoring.}
\label{sec:appendix-normalized-accuracy}
We lowercase each extracted answer and its target, split them on commas, trim whitespace around each token, discard empty tokens, and map \texttt{diamond} to \texttt{rhombus}. A prediction is correct only if the resulting token sequences match exactly, including order and repetitions. Additional nonempty content counts as an error. We report accuracy as the percentage of correct predictions.

\paragraph{Checkpoints and hardware.}
The six open-weight models of Section~\ref{sec:main-results} are the instruction-tuned checkpoints \texttt{Qwen/Qwen3.5-4B}, \texttt{Qwen/Qwen3.5-27B}, and \texttt{Qwen/Qwen3.5-35B-A3B} from the Qwen3.5 family and \texttt{google/gemma-4-E4B-it}, \texttt{google/gemma-4-26B-A4B-it}, and \texttt{google/gemma-4-31B-it} from the Gemma~4 family. All six models run locally with Hugging Face \texttt{transformers} in \texttt{bfloat16} using each model's own chat template, on a single NVIDIA H100 GPU per run. GPT-5.4 (version \texttt{gpt-5.4-2026-03-05}) is accessed through the OpenAI Responses API, with images supplied as base64-encoded JPEGs.

\paragraph{Main benchmark decoding.}
We keep decoding settings fixed across modalities. The main benchmark uses greedy decoding with a 32-token output budget, so no sampling temperature applies. The random seeds 42, 1004, and 1126 control demonstration sampling and the support--query split. Evaluation coverage is documented in Appendix~\ref{sec:evaluation-coverage}.

\paragraph{Diagnostic experiment setup.}
\label{sec:appendix-diagnostic-settings}
The intervention ablations in Section~\ref{sec:closing-gap} and the
instruction and context controls in Section~\ref{sec:instruction-context-controls}
evaluate GPT-5.4 and Qwen3.5-35B-A3B on the same five-task diagnostic
subset, listed in Tables~\ref{tab:intervention-details-gpt}
and~\ref{tab:intervention-details-qwen}.
Each task and condition uses 100 queries, with one demonstration-sampling
seed and a fixed rendering variant within each modality.
Text and image conditions use paired renderings of the same underlying
query inputs and, where present, demonstration inputs. Within the instruction and context controls, the query is
held fixed across prompt settings; the sampled demonstration examples
are also held fixed wherever they are included.
The intervention ablations use 16 input--output demonstrations.
The instruction and context controls vary whether the query appears
alone, with 16 demonstration inputs without their outputs, or with
16 complete input--output demonstrations, as specified in
Table~\ref{tab:instruction-context-setups}.
Summary accuracies give the five tasks equal weight and use the
normalized scoring procedure above.
Appendix~\ref{sec:prompt-templates} provides the prompt formats and
ground-truth task instructions.

\paragraph{Diagnostic decoding settings.}
Thinking conditions reuse the same checkpoints with the chat template's thinking mode enabled. Qwen3.5-35B-A3B thinking uses the recommended sampling settings (temperature 1.0, top-$p$ 0.95, top-$k$ 20, presence penalty 1.5) with a 32{,}768-token thinking budget and a 256-token answer budget, and the corresponding non-thinking comparisons in Sections~\ref{sec:closing-gap}--\ref{sec:instruction-context-controls} use the same sampling settings with a 256-token output budget. GPT-5.4 uses reasoning effort \texttt{high} in the thinking condition, and effort \texttt{none} with a 256-token output cap otherwise, at the API's default temperature and top-$p$ of 1.0.

\FloatBarrier
\section{Full Task Taxonomy}
\label{sec:appendix-task-taxonomy}

Table~\ref{tab:task-taxonomy} lists the full names and definitions of the 38
tasks, grouped by family. Group names describe the shared operation;
the full names of position-controlled tasks additionally specify the positions
and sequence length.
For example, the adjacent swap group contains ``Swap positions 1 and 2 of 7.''
Results tables use abbreviated task labels, defined in their captions.

\begin{table*}[!tbp]
\centering
\scriptsize
\setlength{\tabcolsep}{4pt}
\renewcommand{\arraystretch}{0.92}
\begin{tabular}{@{}p{0.12\textwidth}p{0.28\textwidth}p{0.54\textwidth}@{}}
\toprule
\textbf{Task family} & \textbf{Concrete task} & \textbf{Description} \\
\midrule
\multirow[t]{7}{*}{Selection}
& Select 1st of 5 & Return the first item in a length-5 sequence. \\
& Select 2nd of 5 & Return the second item in a length-5 sequence. \\
& Select 3rd of 5 & Return the third item in a length-5 sequence. \\
& Select 4th of 5 & Return the fourth item in a length-5 sequence. \\
& Select 5th of 5 & Return the fifth item in a length-5 sequence. \\
& Select most frequent of 9 & Return the uniquely most frequent item in a length-9 sequence. \\
& Select most frequent of 11 & Return the uniquely most frequent item in a length-11 sequence. \\
\midrule
\multirow[t]{10}{*}{Relation}
& Even/odd sequence length & Return \texttt{foo} for an even length and \texttt{bar} for an odd length. \\
& Same/different at positions 1 and 2 of 2 & Test whether the two items in a length-2 sequence match. \\
& Same/different at positions 1 and 5 of 5 & Test whether the first and last items in a length-5 sequence match. \\
& Same/different at positions 1 and 7 of 7 & Test whether the first and last items in a length-7 sequence match. \\
& Same/different at positions 1 and 2 of 7 & Test whether the items at positions 1 and 2 match. \\
& Same/different at positions 2 and 3 of 7 & Test whether the items at positions 2 and 3 match. \\
& Same/different at positions 3 and 4 of 7 & Test whether the items at positions 3 and 4 match. \\
& Same/different at positions 4 and 5 of 7 & Test whether the items at positions 4 and 5 match. \\
& Same/different at positions 5 and 6 of 7 & Test whether the items at positions 5 and 6 match. \\
& Same/different at positions 6 and 7 of 7 & Test whether the items at positions 6 and 7 match. \\
\midrule
\multirow[t]{2}{*}{Aggregation}
& Count sequence length & Return the number of items in the sequence. \\
& Count unique items & Return the number of distinct items in the sequence. \\
\midrule
\multirow[t]{19}{*}{Transformation}
& Merge adjacent duplicates in 5 & Collapse each run of identical neighboring items in a length-5 sequence. \\
& Merge adjacent duplicates in 7 & Collapse each run of identical neighboring items in a length-7 sequence. \\
& Reverse 5 & Reverse the order of a length-5 sequence. \\
& Reverse 7 & Reverse the order of a length-7 sequence. \\
& Rotate left by 2 in 5 & Cyclically shift a length-5 sequence two positions to the left. \\
& Rotate left by 2 in 7 & Cyclically shift a length-7 sequence two positions to the left. \\
& Rotate right by 2 in 5 & Cyclically shift a length-5 sequence two positions to the right. \\
& Rotate right by 2 in 7 & Cyclically shift a length-7 sequence two positions to the right. \\
& Reverse window 1--3 of 7 & Reverse positions 1 through 3 and leave all other positions unchanged. \\
& Reverse window 2--4 of 7 & Reverse positions 2 through 4 and leave all other positions unchanged. \\
& Reverse window 3--5 of 7 & Reverse positions 3 through 5 and leave all other positions unchanged. \\
& Reverse window 4--6 of 7 & Reverse positions 4 through 6 and leave all other positions unchanged. \\
& Reverse window 5--7 of 7 & Reverse positions 5 through 7 and leave all other positions unchanged. \\
& Swap positions 1 and 2 of 7 & Swap the items at positions 1 and 2. \\
& Swap positions 2 and 3 of 7 & Swap the items at positions 2 and 3. \\
& Swap positions 3 and 4 of 7 & Swap the items at positions 3 and 4. \\
& Swap positions 4 and 5 of 7 & Swap the items at positions 4 and 5. \\
& Swap positions 5 and 6 of 7 & Swap the items at positions 5 and 6. \\
& Swap positions 6 and 7 of 7 & Swap the items at positions 6 and 7. \\
\bottomrule
\end{tabular}
\caption{The 38 concrete tasks in \benchmarkname{}, grouped by task family. Same/different tasks return \texttt{foo} when the compared items match and \texttt{bar} otherwise.}
\label{tab:task-taxonomy}
\end{table*}

\FloatBarrier
\section{Complete Intervention Results}
    \label{sec:appendix-intervention-details}
    
    Tables~\ref{tab:intervention-details-gpt}
    and~\ref{tab:intervention-details-qwen}
    report all intervention combinations, including those omitted
    from the cumulative path in Table~\ref{tab:prompting-results}.
    Each table gives task-level accuracies and their mean, with
    separate panels for text and image inputs.
    The protocol uses 16 demonstrations, 100 queries per task and
    condition, one demonstration-sampling seed, and one rendering
    variant per modality, as described in Section~\ref{sec:closing-gap}.
    These results support the individual and combined comparisons
    in Section~\ref{sec:closing-gap}.

    \begin{table*}[!tbp]
    \centering
    \small
    \setlength{\tabcolsep}{3pt}
    \renewcommand{\arraystretch}{1.04}

    \begin{tabular*}{\textwidth}{
        @{\extracolsep{\fill}}
        ccc rrrrr r
        @{}
    }
        \toprule
        \multicolumn{3}{c}{\textbf{Interventions}}
            & \multicolumn{6}{c}{\textbf{GPT-5.4}: accuracy (\%)} \\
        \cmidrule(lr){1-3}
        \cmidrule(lr){4-9}
        \textbf{Caption} & \textbf{Meta} & \textbf{Thinking}
            & \shortstack{Most frequent\\(11)}
            & \shortstack{Merge duplicates\\(7)}
            & \shortstack{Reverse\\(7)}
            & \shortstack{Rotate left by 2\\(7)}
            & \shortstack{Select 2nd\\of 5}
            & \textbf{Mean} \\
        \midrule
       
        \multicolumn{9}{l}{\textit{Text inputs}} \\
        
        -- & -- & --
            & 100.0 &  99.0 & 100.0 &  13.0 &  27.0
            & \textbf{67.8} \\
       
        -- & $\checkmark$ & --
            & 100.0 & 100.0 & 100.0 & 100.0 &  21.0
            & \textbf{84.2} \\
        
        -- & -- & $\checkmark$
            &  96.0 &  52.0 &  65.0 &  72.0 &  69.0
            & \textbf{70.8} \\
        
        -- & $\checkmark$ & $\checkmark$
            & 100.0 & 100.0 & 100.0 & 100.0 & 100.0
            & \textbf{100.0} \\
        \addlinespace[4pt]
        
        \multicolumn{9}{l}{\textit{Image inputs}} \\
        
        -- & -- & --
            &  80.0 &  82.0 &  96.0 &  19.0 &  20.0
            & \textbf{59.4} \\
        \addlinespace[2pt]
        
        $\checkmark$ & -- & --
            &  98.0 &  96.0 & 100.0 &  31.0 &  46.0
            & \textbf{74.2} \\
        
        -- & $\checkmark$ & --
            &  85.0 &  72.0 &  98.0 &  85.0 &  28.0
            & \textbf{73.6} \\
        
        -- & -- & $\checkmark$
            &  98.0 &  98.0 &  97.0 & 100.0 & 100.0
            & \textbf{98.6} \\
        \addlinespace[2pt]
        
        $\checkmark$ & $\checkmark$ & --
            & 100.0 &  98.0 & 100.0 &  94.0 &  27.0
            & \textbf{83.8} \\
        
        $\checkmark$ & -- & $\checkmark$
            & 100.0 & 100.0 & 100.0 & 100.0 &  99.0
            & \textbf{99.8} \\
        
        -- & $\checkmark$ & $\checkmark$
            &  99.0 &  98.0 &  99.0 &  99.0 & 100.0
            & \textbf{99.0} \\
        \addlinespace[2pt]
        $\checkmark$ & $\checkmark$ & $\checkmark$
            & 100.0 & 100.0 & 100.0 & 100.0 & 100.0
            & \textbf{100.0} \\
        \bottomrule
    \end{tabular*}

    \caption{
        Task-level intervention accuracy (\%) for GPT-5.4 at 16 shots.
        Checks indicate enabled interventions.
        The image panel contains all eight intervention combinations;
        rows are grouped by the number of enabled interventions,
        not a cumulative sequence.
        Captions apply only to image inputs.
        Each image condition is compared with the text row having
        the same meta-instruction and thinking settings.
        Numbers in task headers give input sequence lengths;
        means weight the five tasks equally.
    }
    \label{tab:intervention-details-gpt}
    \end{table*}

    \begin{table*}[!tbp]
    \centering
    \small
    \setlength{\tabcolsep}{3pt}
    \renewcommand{\arraystretch}{1.04}

    \begin{tabular*}{\textwidth}{
        @{\extracolsep{\fill}}
        ccc rrrrr r
        @{}
    }
        \toprule
        \multicolumn{3}{c}{\textbf{Interventions}}
            & \multicolumn{6}{c}{
                \textbf{Qwen3.5-35B-A3B}: accuracy (\%)
            } \\
        \cmidrule(lr){1-3}
        \cmidrule(lr){4-9}
        \textbf{Caption} & \textbf{Meta} & \textbf{Thinking}
            & \shortstack{Most frequent\\(11)}
            & \shortstack{Merge duplicates\\(7)}
            & \shortstack{Reverse\\(7)}
            & \shortstack{Rotate left by 2\\(7)}
            & \shortstack{Select 2nd\\of 5}
            & \textbf{Mean} \\
        \midrule
        
        \multicolumn{9}{l}{\textit{Text inputs}} \\
        
        -- & -- & --
            & 96.0 & 54.0 & 95.0 & 28.0 & 51.0
            & \textbf{64.8} \\
        
        -- & $\checkmark$ & --
            & 95.0 & 51.0 & 97.0 & 26.0 & 60.0
            & \textbf{65.8} \\
        
        -- & -- & $\checkmark$
            & 99.0 & 97.0 & 96.0 & 97.0 & 83.0
            & \textbf{94.4} \\
        
        -- & $\checkmark$ & $\checkmark$
            & 99.0 & 94.0 & 87.0 & 96.0 & 97.0
            & \textbf{94.6} \\
        \addlinespace[4pt]
        
        \multicolumn{9}{l}{\textit{Image inputs}} \\
        
        -- & -- & --
            & 17.0 & 46.0 &  93.0 &  4.0 & 31.0
            & \textbf{38.2} \\
        \addlinespace[2pt]
        
        $\checkmark$ & -- & --
            & 47.0 & 58.0 & 100.0 & 28.0 & 76.0
            & \textbf{61.8} \\
        
        -- & $\checkmark$ & --
            & 16.0 & 34.0 &  89.0 &  6.0 & 26.0
            & \textbf{34.2} \\
        
        -- & -- & $\checkmark$
            & 17.0 & 11.0 &  39.0 & 21.0 & 45.0
            & \textbf{26.6} \\
        \addlinespace[2pt]
        
        $\checkmark$ & $\checkmark$ & --
            & 59.0 & 77.0 & 100.0 & 45.0 & 81.0
            & \textbf{72.4} \\
        
        $\checkmark$ & -- & $\checkmark$
            & 62.0 & 67.0 &  99.0 & 79.0 & 97.0
            & \textbf{80.8} \\
        
        -- & $\checkmark$ & $\checkmark$
            & 77.0 & 32.0 &  62.0 & 55.0 & 83.0
            & \textbf{61.8} \\
        \addlinespace[2pt]
        
        $\checkmark$ & $\checkmark$ & $\checkmark$
            & 96.0 & 91.0 & 100.0 & 99.0 & 99.0
            & \textbf{97.0} \\
        \bottomrule
    \end{tabular*}

    \caption{
        Task-level intervention accuracy (\%) for Qwen3.5-35B-A3B at 16 shots.
        Checks indicate enabled interventions.
        The image panel contains all eight intervention combinations;
        rows are grouped by the number of enabled interventions,
        not a cumulative sequence.
        Captions apply only to image inputs.
        Each image condition is compared with the text row having
        the same meta-instruction and thinking settings.
        Numbers in task headers give input sequence lengths;
        means weight the five tasks equally.
    }
    \label{tab:intervention-details-qwen}
\end{table*}

\FloatBarrier
\section{Instruction and Context Controls with Thinking}
\label{sec:appendix-instruction-context-thinking}

Table~\ref{tab:instruction-context-controls-thinking} reports the task-level counterpart of Table~\ref{tab:instruction-context-controls} when thinking is enabled. We keep this second reasoning regime separate from the main table so that the primary comparison varies only the instruction and context, while still exposing the model- and task-specific interactions with thinking.
Each task and condition uses 100 queries, one demonstration-sampling
seed, and one rendering variant per modality, following
Section~\ref{sec:instruction-context-controls}; thinking settings are
specified in Appendix~\ref{sec:appendix-eval-details}.

\begin{table*}[!tbp]
\centering
\scriptsize
\setlength{\tabcolsep}{2.4pt}
\renewcommand{\arraystretch}{1.08}
\resizebox{\textwidth}{!}{%
\begin{tabular}{l rrr rrr rrr rrr}
\toprule
\multirow{2}{*}{\textbf{Task}}
& \multicolumn{3}{c}{\textbf{0-shot + GT}}
& \multicolumn{3}{c}{\textbf{Multi-input + GT}}
& \multicolumn{3}{c}{\textbf{16-shot + GT}}
& \multicolumn{3}{c}{\textbf{Regular 16-shot}} \\
\cmidrule(lr){2-4}
\cmidrule(lr){5-7}
\cmidrule(lr){8-10}
\cmidrule(lr){11-13}
& \textbf{T} & \textbf{I} & \textbf{Gap}
& \textbf{T} & \textbf{I} & \textbf{Gap}
& \textbf{T} & \textbf{I} & \textbf{Gap}
& \textbf{T} & \textbf{I} & \textbf{Gap} \\
\midrule
\multicolumn{13}{l}{\textit{GPT-5.4}} \\
Select most frequent of 11
& 100.0 & 99.0 & 1.0 & 100.0 & 86.0 & \textbf{14.0} & 100.0 & 97.0 & 3.0 & 96.0 & 98.0 & $-$2.0 \\
Merge adjacent duplicates in 7
& 100.0 & 92.0 & 8.0 & 100.0 & 68.0 & \textbf{32.0} & 100.0 & 99.0 & 1.0 & 52.0 & 98.0 & $-$46.0 \\
Reverse 7
& 100.0 & 92.0 & 8.0 & 100.0 & 90.0 & \textbf{10.0} & 100.0 & 93.0 & 7.0 & 65.0 & 97.0 & $-$32.0 \\
Rotate left by 2 in 7
& 100.0 & 91.0 & 9.0 & 100.0 & 82.0 & \textbf{18.0} & 100.0 & 93.0 & 7.0 & 72.0 & 100.0 & $-$28.0 \\
Select 2nd of 5
& 100.0 & 97.0 & 3.0 & 100.0 & 96.0 & \textbf{4.0} & 100.0 & 100.0 & 0.0 & 69.0 & 100.0 & $-$31.0 \\
\midrule
\textbf{Average}
& \textbf{100.0} & \textbf{94.2} & \textbf{5.8} & \textbf{100.0} & \textbf{84.4} & \textbf{15.6} & \textbf{100.0} & \textbf{96.4} & \textbf{3.6} & \textbf{70.8} & \textbf{98.6} & \textbf{$-$27.8} \\
\midrule
\multicolumn{13}{l}{\textit{Qwen3.5-35B-A3B}} \\
Select most frequent of 11
& 100.0 & 81.0 & 19.0 & 99.0 & 82.0 & \textbf{17.0} & 100.0 & 82.0 & 18.0 & 99.0 & 17.0 & 82.0 \\
Merge adjacent duplicates in 7
& 100.0 & 42.0 & 58.0 & 100.0 & 31.0 & \textbf{69.0} & 100.0 & 78.0 & 22.0 & 97.0 & 11.0 & 86.0 \\
Reverse 7
& 100.0 & 40.0 & 60.0 & 100.0 & 38.0 & \textbf{62.0} & 100.0 & 79.0 & 21.0 & 96.0 & 39.0 & 57.0 \\
Rotate left by 2 in 7
& 100.0 & 33.0 & 67.0 & 100.0 & 34.0 & \textbf{66.0} & 100.0 & 75.0 & 25.0 & 97.0 & 21.0 & 76.0 \\
Select 2nd of 5
& 100.0 & 88.0 & 12.0 & 100.0 & 85.0 & \textbf{15.0} & 100.0 & 96.0 & 4.0 & 83.0 & 45.0 & 38.0 \\
\midrule
\textbf{Average}
& \textbf{100.0} & \textbf{56.8} & \textbf{43.2} & \textbf{99.8} & \textbf{54.0} & \textbf{45.8} & \textbf{100.0} & \textbf{82.0} & \textbf{18.0} & \textbf{94.4} & \textbf{26.6} & \textbf{67.8} \\
\bottomrule
\end{tabular}%
}
\caption{Instruction and context controls with thinking enabled on five diagnostic tasks. \textbf{T} and \textbf{I} report accuracy (\%) in the text and image conditions, respectively; \textbf{Gap} is T $-$ I within the same row and condition, in percentage points. Positive gaps indicate lower accuracy in the image condition. GT denotes a ground-truth task instruction; Table~\ref{tab:instruction-context-setups} defines the conditions. Average rows give each task equal weight. Boldface highlights the multi-input gaps.}
\label{tab:instruction-context-controls-thinking}
\end{table*}

\FloatBarrier
\section{Detailed Benchmark Results}
\label{sec:additional-general-inference-tables}

\FloatBarrier
\subsection{Complete Task-Level Results}

Tables~\ref{tab:full-task-16shot-gap-selection}--\ref{tab:full-task-16shot-gap-transformation-swaps} report each model's accuracy on all 38 tasks at 16 shots, allowing readers to inspect task-specific differences that the family averages in Table~\ref{tab:main-benchmark-results} summarize.

\begin{table*}[!tbp]
\scriptsize
\setlength{\tabcolsep}{3pt}
\renewcommand{\arraystretch}{1.08}
\centering
\begin{tabular}{lcccccccccccccc}
\toprule
\multirow{3}{*}{\textbf{Model}} & \multicolumn{14}{c}{\textbf{Selection}} \\
\cmidrule(lr){2-15}
 & \multicolumn{2}{c}{\rotatebox{60}{\makebox[1.35cm][l]{Select 1/5}}} & \multicolumn{2}{c}{\rotatebox{60}{\makebox[1.35cm][l]{Select 2/5}}} & \multicolumn{2}{c}{\rotatebox{60}{\makebox[1.35cm][l]{Select 3/5}}} & \multicolumn{2}{c}{\rotatebox{60}{\makebox[1.35cm][l]{Select 4/5}}} & \multicolumn{2}{c}{\rotatebox{60}{\makebox[1.35cm][l]{Select 5/5}}} & \multicolumn{2}{c}{\rotatebox{60}{\makebox[1.35cm][l]{Freq. 9}}} & \multicolumn{2}{c}{\rotatebox{60}{\makebox[1.35cm][l]{Freq. 11}}} \\
\cmidrule(lr){2-3}
\cmidrule(lr){4-5}
\cmidrule(lr){6-7}
\cmidrule(lr){8-9}
\cmidrule(lr){10-11}
\cmidrule(lr){12-13}
\cmidrule(lr){14-15}
 & \textbf{T} & \textbf{I} & \textbf{T} & \textbf{I} & \textbf{T} & \textbf{I} & \textbf{T} & \textbf{I} & \textbf{T} & \textbf{I} & \textbf{T} & \textbf{I} & \textbf{T} & \textbf{I} \\
\midrule
Qwen3.5-35B-A3B & 100.0 & 90.2 & 52.9 & 36.9 & 31.0 & 25.3 & 25.4 & 32.8 & 49.2 & 71.2 & 99.6 & 37.6 & 98.5 & 40.2 \\
Qwen3.5-27B & 100.0 & 91.1 & 64.2 & 33.5 & 31.0 & 26.2 & 25.5 & 21.4 & 98.7 & 81.7 & 100.0 & 59.3 & 99.9 & 48.0 \\
Qwen3.5-4B & 100.0 & 86.6 & 39.6 & 18.8 & 26.0 & 13.5 & 19.2 & 11.1 & 57.4 & 13.8 & 92.6 & 27.7 & 89.1 & 38.3 \\
Gemma 4-31B IT & 100.0 & 98.9 & 99.6 & 23.4 & 92.5 & 52.4 & 73.8 & 27.4 & 100.0 & 99.5 & 99.9 & 58.7 & 99.8 & 60.7 \\
Gemma 4-26B-A4B IT & 99.7 & 92.8 & 95.7 & 9.2 & 75.9 & 10.8 & 55.7 & 15.0 & 99.9 & 91.7 & 99.9 & 30.9 & 99.5 & 30.1 \\
Gemma 4-E4B IT & 100.0 & 67.2 & 36.6 & 31.7 & 28.3 & 29.8 & 26.4 & 29.4 & 94.7 & 49.8 & 99.7 & 15.8 & 97.0 & 13.5 \\
\bottomrule
\end{tabular}%
\caption{Complete 16-shot task-level benchmark results for Selection. Select $k/5$ returns the shape at position $k$ in a five-shape sequence; Freq. $n$ selects the most frequent shape in an $n$-shape sequence. Rows are models and columns are concrete tasks; \textbf{T} and \textbf{I} report accuracy (\%) in the text and image conditions, respectively. Each value averages available seeds and rendering variants within the corresponding task.}
\label{tab:full-task-16shot-gap-selection}
\end{table*}

\begin{table*}[!tbp]
\tiny
\setlength{\tabcolsep}{2pt}
\renewcommand{\arraystretch}{1.08}
\centering
\resizebox{\linewidth}{!}{%
\begin{tabular}{lcccccccccccccccccccc}
\toprule
\multirow{3}{*}{\textbf{Model}} & \multicolumn{20}{c}{\textbf{Relation}} \\
\cmidrule(lr){2-21}
 & \multicolumn{2}{c}{\rotatebox{60}{\makebox[1.35cm][l]{Even/odd}}} & \multicolumn{2}{c}{\rotatebox{60}{\makebox[1.35cm][l]{Same/diff. 2}}} & \multicolumn{2}{c}{\rotatebox{60}{\makebox[1.35cm][l]{Same/diff. 5}}} & \multicolumn{2}{c}{\rotatebox{60}{\makebox[1.35cm][l]{Same/diff. 7}}} & \multicolumn{2}{c}{\rotatebox{60}{\makebox[1.35cm][l]{Adj. same/diff. 1/7}}} & \multicolumn{2}{c}{\rotatebox{60}{\makebox[1.35cm][l]{Adj. same/diff. 2/7}}} & \multicolumn{2}{c}{\rotatebox{60}{\makebox[1.35cm][l]{Adj. same/diff. 3/7}}} & \multicolumn{2}{c}{\rotatebox{60}{\makebox[1.35cm][l]{Adj. same/diff. 4/7}}} & \multicolumn{2}{c}{\rotatebox{60}{\makebox[1.35cm][l]{Adj. same/diff. 5/7}}} & \multicolumn{2}{c}{\rotatebox{60}{\makebox[1.35cm][l]{Adj. same/diff. 6/7}}} \\
\cmidrule(lr){2-3}
\cmidrule(lr){4-5}
\cmidrule(lr){6-7}
\cmidrule(lr){8-9}
\cmidrule(lr){10-11}
\cmidrule(lr){12-13}
\cmidrule(lr){14-15}
\cmidrule(lr){16-17}
\cmidrule(lr){18-19}
\cmidrule(lr){20-21}
 & \textbf{T} & \textbf{I} & \textbf{T} & \textbf{I} & \textbf{T} & \textbf{I} & \textbf{T} & \textbf{I} & \textbf{T} & \textbf{I} & \textbf{T} & \textbf{I} & \textbf{T} & \textbf{I} & \textbf{T} & \textbf{I} & \textbf{T} & \textbf{I} & \textbf{T} & \textbf{I} \\
\midrule
Qwen3.5-35B-A3B & 46.1 & 43.8 & 99.0 & 84.4 & 87.9 & 79.3 & 87.8 & 74.8 & 57.0 & 51.7 & 50.7 & 51.3 & 51.0 & 52.0 & 50.7 & 50.7 & 46.0 & 52.0 & 53.7 & 46.7 \\
Qwen3.5-27B & 49.8 & 45.4 & 99.6 & 93.2 & 88.8 & 88.7 & 85.1 & 88.0 & 52.0 & 50.0 & 48.7 & 48.3 & 51.3 & 49.0 & 51.3 & 49.7 & 45.0 & 51.0 & 51.7 & 47.0 \\
Qwen3.5-4B & 49.1 & 54.7 & 98.4 & 90.8 & 88.7 & 88.7 & 88.0 & 88.0 & 49.7 & 51.0 & 49.3 & 51.0 & 50.3 & 48.3 & 49.3 & 51.7 & 50.7 & 52.0 & 43.3 & 49.0 \\
Gemma 4-31B IT & 49.8 & 49.8 & 100.0 & 93.2 & 85.8 & 86.8 & 77.2 & 78.3 & 66.7 & 50.3 & 55.3 & 53.3 & 53.3 & 50.7 & 45.3 & 52.7 & 56.0 & 50.0 & 60.3 & 51.0 \\
Gemma 4-26B-A4B IT & 49.0 & 50.3 & 98.7 & 88.7 & 87.6 & 80.9 & 86.8 & 87.2 & 51.0 & 53.0 & 51.3 & 49.0 & 48.0 & 50.3 & 54.3 & 51.3 & 51.3 & 49.0 & 53.7 & 51.0 \\
Gemma 4-E4B IT & 51.3 & 46.6 & 97.7 & 90.8 & 87.4 & 88.7 & 87.8 & 88.0 & 51.7 & 48.7 & 53.3 & 50.7 & 50.3 & 49.3 & 51.0 & 53.7 & 47.3 & 48.0 & 54.3 & 47.3 \\
\bottomrule
\end{tabular}%
}
\caption{Complete 16-shot task-level benchmark results for Relation. Same/diff. $n$ compares the first and last shapes in an $n$-shape sequence. Adj. same/diff. denotes adjacent same/different; $k/7$ identifies the pair at positions $k$ and $k+1$ in a seven-shape sequence. Rows are models and columns are concrete tasks; \textbf{T} and \textbf{I} report accuracy (\%) in the text and image conditions, respectively. Each value averages available seeds and rendering variants within the corresponding task.}
\label{tab:full-task-16shot-gap-relation}
\end{table*}

\begin{table*}[!tbp]
\scriptsize
\setlength{\tabcolsep}{3pt}
\renewcommand{\arraystretch}{1.08}
\centering
\begin{tabular}{lcccc}
\toprule
\multirow{3}{*}{\textbf{Model}} & \multicolumn{4}{c}{\textbf{Aggregation}} \\
\cmidrule(lr){2-5}
 & \multicolumn{2}{c}{\rotatebox{60}{\makebox[1.35cm][l]{Count}}} & \multicolumn{2}{c}{\rotatebox{60}{\makebox[1.35cm][l]{Count uniq.}}} \\
\cmidrule(lr){2-3}
\cmidrule(lr){4-5}
 & \textbf{T} & \textbf{I} & \textbf{T} & \textbf{I} \\
\midrule
Qwen3.5-35B-A3B & 100.0 & 99.5 & 68.9 & 43.5 \\
Qwen3.5-27B & 100.0 & 100.0 & 79.7 & 46.7 \\
Qwen3.5-4B & 99.8 & 95.8 & 43.0 & 43.9 \\
Gemma 4-31B IT & 100.0 & 93.8 & 90.6 & 43.9 \\
Gemma 4-26B-A4B IT & 100.0 & 92.9 & 63.2 & 42.0 \\
Gemma 4-E4B IT & 99.0 & 57.9 & 49.0 & 41.1 \\
\bottomrule
\end{tabular}%
\caption{Complete 16-shot task-level benchmark results for Aggregation. Rows are models and columns are concrete tasks; \textbf{T} and \textbf{I} report accuracy (\%) in the text and image conditions, respectively. Each value averages available seeds and rendering variants within the corresponding task.}
\label{tab:full-task-16shot-gap-aggregation}
\end{table*}

\begin{table*}[!tbp]
\centering
\scriptsize
\setlength{\tabcolsep}{3pt}
\renewcommand{\arraystretch}{1.08}
\begin{tabular}{lcccccccccccccccc}
\toprule
\multirow{3}{*}{\textbf{Model}} & \multicolumn{16}{c}{\textbf{Transformation: merge, full reversal, and rotation}} \\
\cmidrule(lr){2-17}
 & \multicolumn{2}{c}{\rotatebox{60}{\makebox[1.35cm][l]{Merge dup. 5}}} & \multicolumn{2}{c}{\rotatebox{60}{\makebox[1.35cm][l]{Merge dup. 7}}} & \multicolumn{2}{c}{\rotatebox{60}{\makebox[1.35cm][l]{Reverse 5}}} & \multicolumn{2}{c}{\rotatebox{60}{\makebox[1.35cm][l]{Reverse 7}}} & \multicolumn{2}{c}{\rotatebox{60}{\makebox[1.35cm][l]{Rot. L2/5}}} & \multicolumn{2}{c}{\rotatebox{60}{\makebox[1.35cm][l]{Rot. L2/7}}} & \multicolumn{2}{c}{\rotatebox{60}{\makebox[1.35cm][l]{Rot. R2/5}}} & \multicolumn{2}{c}{\rotatebox{60}{\makebox[1.35cm][l]{Rot. R2/7}}} \\
\cmidrule(lr){2-3}
\cmidrule(lr){4-5}
\cmidrule(lr){6-7}
\cmidrule(lr){8-9}
\cmidrule(lr){10-11}
\cmidrule(lr){12-13}
\cmidrule(lr){14-15}
\cmidrule(lr){16-17}
 & \textbf{T} & \textbf{I} & \textbf{T} & \textbf{I} & \textbf{T} & \textbf{I} & \textbf{T} & \textbf{I} & \textbf{T} & \textbf{I} & \textbf{T} & \textbf{I} & \textbf{T} & \textbf{I} & \textbf{T} & \textbf{I} \\
\midrule
Qwen3.5-35B-A3B & 79.7 & 72.2 & 43.7 & 39.8 & 94.1 & 89.9 & 98.4 & 91.1 & 27.1 & 3.5 & 39.2 & 5.9 & 13.4 & 4.2 & 15.8 & 2.2 \\
Qwen3.5-27B & 91.8 & 79.6 & 87.2 & 68.3 & 100.0 & 97.2 & 100.0 & 93.2 & 50.4 & 6.4 & 65.2 & 6.3 & 49.3 & 5.7 & 30.4 & 5.6 \\
Qwen3.5-4B & 84.7 & 58.8 & 70.8 & 41.7 & 50.6 & 29.4 & 62.3 & 17.2 & 3.5 & 0.5 & 13.2 & 0.0 & 7.8 & 0.0 & 10.7 & 0.0 \\
Gemma 4-31B IT & 91.4 & 73.1 & 99.2 & 31.8 & 100.0 & 95.0 & 100.0 & 86.8 & 44.0 & 4.8 & 46.1 & 3.0 & 75.9 & 2.2 & 89.8 & 4.6 \\
Gemma 4-26B-A4B IT & 86.5 & 39.7 & 67.3 & 7.3 & 99.1 & 65.6 & 99.6 & 9.5 & 24.6 & 0.0 & 39.5 & 0.0 & 15.0 & 0.1 & 18.5 & 0.0 \\
Gemma 4-E4B IT & 83.4 & 23.3 & 52.0 & 3.6 & 28.0 & 2.4 & 26.4 & 0.0 & 4.1 & 0.2 & 3.9 & 0.0 & 5.8 & 0.0 & 1.2 & 0.0 \\
\bottomrule
\end{tabular}
\caption{Complete 16-shot task-level benchmark results for Transformation: merge, full reversal, and rotation. Rows are models and columns are concrete tasks; \textbf{T} and \textbf{I} report accuracy (\%) in the text and image conditions, respectively. Each value averages available seeds and rendering variants within the corresponding task.}
\label{tab:full-task-16shot-gap-transformation}
\end{table*}

\begin{table*}[!tbp]
\centering
\scriptsize
\setlength{\tabcolsep}{3pt}
\renewcommand{\arraystretch}{1.08}
\begin{tabular}{lcccccccccc}
\toprule
\multirow{3}{*}{\textbf{Model}} & \multicolumn{10}{c}{\textbf{Transformation: window reversals}} \\
\cmidrule(lr){2-11}
 & \multicolumn{2}{c}{\rotatebox{60}{\makebox[1.35cm][l]{Rev. win. 1/7}}} & \multicolumn{2}{c}{\rotatebox{60}{\makebox[1.35cm][l]{Rev. win. 2/7}}} & \multicolumn{2}{c}{\rotatebox{60}{\makebox[1.35cm][l]{Rev. win. 3/7}}} & \multicolumn{2}{c}{\rotatebox{60}{\makebox[1.35cm][l]{Rev. win. 4/7}}} & \multicolumn{2}{c}{\rotatebox{60}{\makebox[1.35cm][l]{Rev. win. 5/7}}} \\
\cmidrule(lr){2-3}
\cmidrule(lr){4-5}
\cmidrule(lr){6-7}
\cmidrule(lr){8-9}
\cmidrule(lr){10-11}
 & \textbf{T} & \textbf{I} & \textbf{T} & \textbf{I} & \textbf{T} & \textbf{I} & \textbf{T} & \textbf{I} & \textbf{T} & \textbf{I} \\
\midrule
Qwen3.5-35B-A3B & 13.3 & 1.7 & 4.0 & 1.7 & 0.7 & 0.0 & 0.3 & 0.0 & 19.0 & 1.3 \\
Qwen3.5-27B & 16.0 & 5.0 & 5.0 & 2.7 & 4.3 & 0.0 & 2.7 & 0.0 & 45.7 & 0.0 \\
Qwen3.5-4B & 4.0 & 0.7 & 1.3 & 0.7 & 1.0 & 0.0 & 0.0 & 0.0 & 0.3 & 0.3 \\
Gemma 4-31B IT & 33.7 & 0.0 & 6.7 & 0.3 & 7.0 & 0.0 & 7.7 & 0.0 & 69.7 & 0.0 \\
Gemma 4-26B-A4B IT & 11.0 & 0.7 & 11.0 & 0.0 & 4.7 & 0.0 & 5.3 & 0.0 & 28.3 & 2.3 \\
Gemma 4-E4B IT & 2.3 & 0.3 & 2.3 & 0.0 & 0.3 & 0.3 & 1.0 & 0.3 & 4.0 & 0.3 \\
\bottomrule
\end{tabular}
\caption{Complete 16-shot task-level benchmark results for Transformation: window reversals. Rev. win. denotes window reversal; $k/7$ identifies the three-shape window starting at position $k$ in a seven-shape sequence. Rows are models and columns are concrete tasks; \textbf{T} and \textbf{I} report accuracy (\%) in the text and image conditions, respectively. Each value averages available seeds and rendering variants within the corresponding task.}
\label{tab:full-task-16shot-gap-transformation-windows}
\end{table*}

\begin{table*}[!tbp]
\centering
\scriptsize
\setlength{\tabcolsep}{3pt}
\renewcommand{\arraystretch}{1.08}
\begin{tabular}{lcccccccccccc}
\toprule
\multirow{3}{*}{\textbf{Model}} & \multicolumn{12}{c}{\textbf{Transformation: adjacent swaps}} \\
\cmidrule(lr){2-13}
 & \multicolumn{2}{c}{\rotatebox{60}{\makebox[1.35cm][l]{Adj. swap 1/7}}} & \multicolumn{2}{c}{\rotatebox{60}{\makebox[1.35cm][l]{Adj. swap 2/7}}} & \multicolumn{2}{c}{\rotatebox{60}{\makebox[1.35cm][l]{Adj. swap 3/7}}} & \multicolumn{2}{c}{\rotatebox{60}{\makebox[1.35cm][l]{Adj. swap 4/7}}} & \multicolumn{2}{c}{\rotatebox{60}{\makebox[1.35cm][l]{Adj. swap 5/7}}} & \multicolumn{2}{c}{\rotatebox{60}{\makebox[1.35cm][l]{Adj. swap 6/7}}} \\
\cmidrule(lr){2-3}
\cmidrule(lr){4-5}
\cmidrule(lr){6-7}
\cmidrule(lr){8-9}
\cmidrule(lr){10-11}
\cmidrule(lr){12-13}
 & \textbf{T} & \textbf{I} & \textbf{T} & \textbf{I} & \textbf{T} & \textbf{I} & \textbf{T} & \textbf{I} & \textbf{T} & \textbf{I} & \textbf{T} & \textbf{I} \\
\midrule
Qwen3.5-35B-A3B & 59.3 & 27.0 & 7.3 & 9.7 & 2.3 & 0.0 & 1.7 & 0.0 & 11.3 & 0.7 & 71.0 & 0.3 \\
Qwen3.5-27B & 100.0 & 23.7 & 50.3 & 16.7 & 36.7 & 2.3 & 30.7 & 1.0 & 28.0 & 0.0 & 99.3 & 0.0 \\
Qwen3.5-4B & 53.7 & 2.0 & 5.7 & 0.0 & 2.3 & 0.0 & 2.7 & 0.0 & 1.7 & 0.0 & 28.3 & 0.7 \\
Gemma 4-31B IT & 78.3 & 1.7 & 42.3 & 0.0 & 41.7 & 0.0 & 28.0 & 0.0 & 33.0 & 0.0 & 95.3 & 0.0 \\
Gemma 4-26B-A4B IT & 78.0 & 0.0 & 25.3 & 0.0 & 31.0 & 0.0 & 43.3 & 0.0 & 41.0 & 0.7 & 88.3 & 1.7 \\
Gemma 4-E4B IT & 44.0 & 1.0 & 8.7 & 0.0 & 1.0 & 0.0 & 2.0 & 0.0 & 4.3 & 0.0 & 13.0 & 0.7 \\
\bottomrule
\end{tabular}
\caption{Complete 16-shot task-level benchmark results for Transformation: adjacent swaps. Adj. swap denotes adjacent swap; $k/7$ identifies the pair at positions $k$ and $k+1$ in a seven-shape sequence. Rows are models and columns are concrete tasks; \textbf{T} and \textbf{I} report accuracy (\%) in the text and image conditions, respectively. Each value averages available seeds and rendering variants within the corresponding task.}
\label{tab:full-task-16shot-gap-transformation-swaps}
\end{table*}

\clearpage
\subsection{Overall Results Across Shot Counts}

Table~\ref{tab:lower-shot-overall-gap} provides the numerical results underlying Figure~\ref{fig:modality-gap}.

\noindent\begin{minipage}[t]{\columnwidth}
\centering
\scriptsize
\setlength{\tabcolsep}{3pt}
\renewcommand{\arraystretch}{1.05}
\begin{tabular}{@{}lrrr@{}}
\toprule
\multicolumn{4}{c}{\textbf{2-shot}} \\
Model & \textbf{T} & \textbf{I} & \textbf{Gap} \\
\midrule
Qwen3.5-35B-A3B & 30.4 & 4.7 & 25.7 \\
Qwen3.5-27B & 40.9 & 21.8 & 19.1 \\
Qwen3.5-4B & 30.4 & 21.9 & 8.5 \\
Gemma 4-31B IT & 37.5 & 26.3 & 11.2 \\
Gemma 4-26B-A4B IT & 35.3 & 24.0 & 11.2 \\
Gemma 4-E4B IT & 29.3 & 21.9 & 7.4 \\
\midrule
\multicolumn{4}{c}{\textbf{4-shot}} \\
Model & \textbf{T} & \textbf{I} & \textbf{Gap} \\
\midrule
Qwen3.5-35B-A3B & 38.3 & 17.0 & 21.3 \\
Qwen3.5-27B & 49.1 & 32.9 & 16.2 \\
Qwen3.5-4B & 35.1 & 25.9 & 9.2 \\
Gemma 4-31B IT & 47.9 & 29.6 & 18.3 \\
Gemma 4-26B-A4B IT & 42.8 & 29.5 & 13.3 \\
Gemma 4-E4B IT & 36.3 & 23.1 & 13.2 \\
\midrule
\multicolumn{4}{c}{\textbf{8-shot}} \\
Model & \textbf{T} & \textbf{I} & \textbf{Gap} \\
\midrule
Qwen3.5-35B-A3B & 44.0 & 31.7 & 12.2 \\
Qwen3.5-27B & 56.0 & 37.0 & 19.1 \\
Qwen3.5-4B & 39.7 & 28.2 & 11.5 \\
Gemma 4-31B IT & 57.9 & 36.5 & 21.3 \\
Gemma 4-26B-A4B IT & 51.5 & 30.8 & 20.7 \\
Gemma 4-E4B IT & 38.9 & 24.5 & 14.3 \\
\midrule
\multicolumn{4}{c}{\textbf{16-shot}} \\
Model & \textbf{T} & \textbf{I} & \textbf{Gap} \\
\midrule
Qwen3.5-35B-A3B & 48.9 & 37.2 & 11.6 \\
Qwen3.5-27B & 60.9 & 40.3 & 20.6 \\
Qwen3.5-4B & 41.8 & 29.6 & 12.1 \\
Gemma 4-31B IT & 68.3 & 38.9 & 29.4 \\
Gemma 4-26B-A4B IT & 58.9 & 30.4 & 28.5 \\
Gemma 4-E4B IT & 40.8 & 25.8 & 15.0 \\
\bottomrule
\end{tabular}

\captionsetup{hypcap=false}
\captionof{table}{ICL results across shot counts, averaged equally over all 38 tasks after averaging available seeds and rendering variants within tasks. Panels give the shot count. \textbf{T} and \textbf{I} report accuracy (\%) in the text and image conditions; \textbf{Gap} is T $-$ I in percentage points. Positive gaps indicate lower accuracy in the image condition. Coverage is documented in Appendix~\ref{sec:evaluation-coverage}.}
\label{tab:lower-shot-overall-gap}
\end{minipage}\par

\newpage
\subsection{Evaluation Coverage}
\label{sec:evaluation-coverage}

Reported results use available runs; additional evaluations are underway
to complete rendering and seed coverage.

Table~\ref{tab:general-inference-coverage} reports the evaluation runs used
for the 38-task benchmark at the four shot counts in
Figure~\ref{fig:modality-gap}. One run evaluates 100 queries for a fixed
model, task, shot count, rendering variant, and seed. The denominators
assume four text variants, four image variants, and three seeds per task.
For reporting current evaluation coverage, the table separates the six adjacent
same/different, five window reversal, and six adjacent swap tasks (17 in total)
from the remaining 21 tasks, which include the five positional selection tasks.
This distinction reflects available runs only; all 38 tasks contribute equally
to the benchmark averages.

\begin{table*}[t]
\centering
\small
\setlength{\tabcolsep}{8pt}
\renewcommand{\arraystretch}{1.05}
\begin{tabular}{@{}lrccc@{}}
\toprule
Model & Shots & \shortstack{Other tasks\\(21)} & \shortstack{Adjacent same/different,\\window reversal, adjacent swap (17)} & \shortstack{Total\\(38 tasks)} \\
\midrule
Qwen3.5-35B-A3B & 2 & 504/504 & 102/408 & 606/912 \\
 & 4 & 504/504 & 102/408 & 606/912 \\
 & 8 & 504/504 & 102/408 & 606/912 \\
 & 16 & 504/504 & 102/408 & 606/912 \\
\midrule
Qwen3.5-27B & 2 & 481/504 & 102/408 & 583/912 \\
 & 4 & 481/504 & 102/408 & 583/912 \\
 & 8 & 481/504 & 102/408 & 583/912 \\
 & 16 & 504/504 & 102/408 & 606/912 \\
\midrule
Qwen3.5-4B & 2 & 337/504 & 102/408 & 439/912 \\
 & 4 & 336/504 & 102/408 & 438/912 \\
 & 8 & 336/504 & 102/408 & 438/912 \\
 & 16 & 504/504 & 102/408 & 606/912 \\
\midrule
Gemma 4-31B IT & 2 & 504/504 & 102/408 & 606/912 \\
 & 4 & 504/504 & 102/408 & 606/912 \\
 & 8 & 504/504 & 102/408 & 606/912 \\
 & 16 & 504/504 & 102/408 & 606/912 \\
\midrule
Gemma 4-26B-A4B IT & 2 & 504/504 & 102/408 & 606/912 \\
 & 4 & 504/504 & 102/408 & 606/912 \\
 & 8 & 504/504 & 102/408 & 606/912 \\
 & 16 & 504/504 & 102/408 & 606/912 \\
\midrule
Gemma 4-E4B IT & 2 & 369/504 & 102/408 & 471/912 \\
 & 4 & 368/504 & 102/408 & 470/912 \\
 & 8 & 367/504 & 102/408 & 469/912 \\
 & 16 & 504/504 & 102/408 & 606/912 \\
\bottomrule
\end{tabular}

\caption{Evaluation coverage underlying the reported benchmark results.
Entries give available/expected runs for the remaining 21 tasks, the 17
adjacent same/different, window reversal, and adjacent swap tasks, and all 38 tasks. Per model and shot count,
the expected counts are $21\times8\times3=504$,
$17\times8\times3=408$, and $38\times8\times3=912$, respectively.
Each of these 17 tasks currently includes one text and one image variant
with all three seeds, giving 102 of 408 runs. The remaining 306 runs per
model and shot count correspond to the other six rendering variants.
For the other 21 tasks, deficits at 2/4/8 shots are exclusively seed-1126 runs:
23/23/23 for Qwen3.5-27B, 167/168/168 for Qwen3.5-4B, and
135/136/137 for Gemma 4-E4B IT. All included variants have three seeds at
16 shots. Only runs included in the reported results are counted.}
\label{tab:general-inference-coverage}
\end{table*}

\end{document}